\documentclass[runningheads]{llncs}

\usepackage{eccv}

\usepackage{eccvabbrv}

\usepackage{graphicx}
\usepackage{booktabs}

\usepackage[accsupp]{axessibility}  

\usepackage{multirow}
\usepackage{hhline}
\usepackage[listings]{tcolorbox}
\usepackage{enumitem}

\definecolor{Gray}{gray}{0.9}
\definecolor{White}{gray}{1}
\definecolor{DGray}{gray}{0.8}
\definecolor{DDDDGray}{gray}{0.3}
\definecolor{WhiteGray}{rgb}{0.9, 0.9, 0.9}
\definecolor{citecolor}{HTML}{0071bc}
\definecolor{darkred}{rgb}{0.6, 0.1, 0.05}
\definecolor{DeltaColor}{rgb}{0.039,0.73,0.71}
\definecolor{SigmaColor}{rgb}{0.98,0.45,0.0}
\definecolor{AlphaColor}{rgb}{0,0,0.8}
\definecolor{BetaColor}{rgb}{0.8,0,0.8}
\definecolor{GammaColor}{rgb}{0.514,0.34,0.224}
\definecolor{EpsilonColor}{rgb}{0.353,0.725,0.906}
\definecolor{PurpleColor}{HTML}{8B008B}
\definecolor{BadColor}{HTML}{C0392B}
\definecolor{OrangeColor}{rgb}{0.914,0.541,0.0.141}
\definecolor{GreenColor}{rgb}{0.137,0.573,0.565}
\definecolor{RedColor}{rgb}{0.949,0.275, 0.224}
\definecolor{LightCyan}{rgb}{0.88,1,1}
\definecolor{Gray}{gray}{0.85}

\definecolor{bestcolor}{rgb}{1, 0.5, 0.25}
\definecolor{secondbestcolor}{rgb}{1, 0.8, 0.5}

\newcommand{\del}[1]{}

\DeclareMathAlphabet\mathbfcal{OMS}{cmsy}{b}{n}

\newcommand\supp{Appx\xspace}

\newcommand{\cheading}[1]{\mbox{\textbf{#1}\;}}

\newcommand{\customfootnotetext}[2]{{
  \renewcommand{\thefootnote}{#1}
  \footnotetext[0]{#2}}}

\newlength\savewidth

\newcommand{\projectURL}{\url{https://kailinli.github.io/SemGrasp}}

\newcommand{\codebracket}[1]{\textless#1\textgreater}
\newcommand{\method}{\textbf{\textsc{SemGrasp}}\xspace}
\newcommand{\dataset}{\textbf{\textsc{CapGrasp}}\xspace}

\newcommand{\grasptoken}{\texttt{\codebracket{grasp}}\xspace}
\newcommand{\mbold}[1]{\boldsymbol{#1}}
\newcommand{\obj}{\mbold{O}}
\newcommand{\hand}{\mbold{H}}
\newcommand{\lang}{\mbold{L}}
\newcommand{\grasp}{\mbold{G}}
\newcommand{\real}{\mathbb{R}}
\newcommand{\tokenO}{\texttt{o}}
\newcommand{\tokenM}{\texttt{m}}
\newcommand{\tokenR}{\texttt{r}}
\newcommand{\tokenZ}{\texttt{z}}

\usepackage[pagebackref,breaklinks,colorlinks,citecolor=eccvblue]{hyperref}

\usepackage{orcidlink}

\begin{document}

\title{\method: Semantic Grasp Generation via Language Aligned Discretization} 

\titlerunning{Semantic Grasp Generation via Language Aligned Discretization}


\author{Kailin Li\inst{1,2} \and
Jingbo Wang\inst{2} \and
Lixin Yang\inst{1} \and
Cewu Lu\inst{1}\textsuperscript{$\boldsymbol{\dagger}$} \and
Bo Dai\inst{2}}

\authorrunning{K.~Li et al.}

\institute{Shanghai Jiao Tong University, Shanghai, China\\
\email{\{kailinli,siriusyang,lucewu\}@sjtu.edu.cn}\\
\and
Shanghai AI Laboratory, Shanghai, China\\
\email{\{wangjingbo,daibo\}@pjlab.org.cn}}

\maketitle

\customfootnotetext{$\dagger$}{
    Corresponding author.
}

\begin{abstract}
        Generating natural human grasps necessitates consideration of not just object geometry but also semantic information. Solely depending on object shape for grasp generation confines the applications of prior methods in downstream tasks. This paper presents a novel semantic-based grasp generation method, termed \textnormal{\method}, which generates a static human grasp pose by incorporating semantic information into the grasp representation. We introduce a discrete representation that aligns the grasp space with semantic space, enabling the generation of grasp postures in accordance with language instructions. A Multimodal Large Language Model (MLLM) is subsequently fine-tuned, integrating object, grasp, and language within a unified semantic space. To facilitate the training of \textnormal{\method}, we have compiled a large-scale, grasp-text-aligned dataset named \textnormal{\dataset}, featuring about 260k detailed captions and 50k diverse grasps. Experimental findings demonstrate that \textnormal{\method} efficiently generates natural human grasps in alignment with linguistic intentions. \del{Our code, models, and dataset will be made publicly available.}Our code, models, and dataset are available publicly at: \projectURL.
\end{abstract}    
\section{Introduction}
\label{sec:intro}
\begin{figure}[t]
    \centering
    \includegraphics[width=\linewidth]{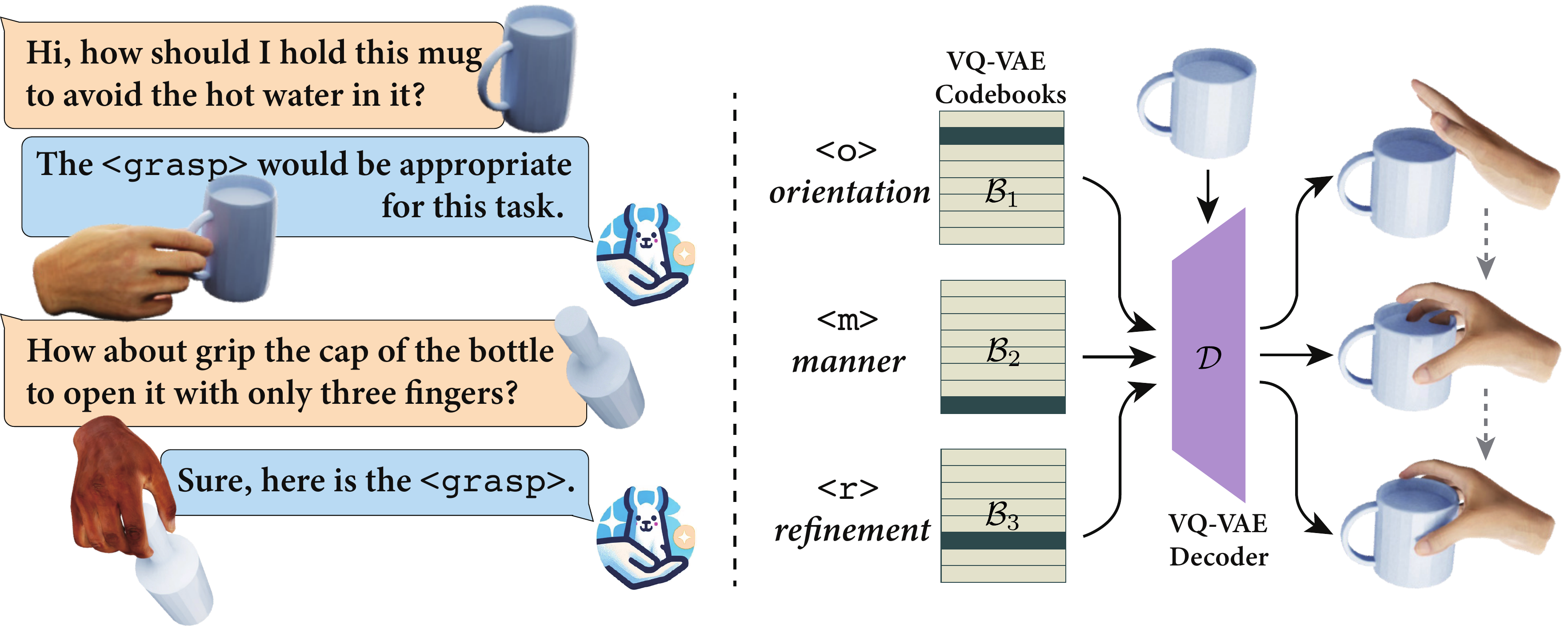}
     \caption{Our \method~methodology. The left figure shows the grasp generation process, while the right illustrates the decoding of our discrete grasp representation.}
     \label{fig:teaser}
  \end{figure}

In applications such as AR/VR and embodied robotics, the ability to generate human-like grasps for a given object is of substantial value. The goal of grasping extends beyond simple object lifting; it involves alignment with human intent and preparation for subsequent manipulation tasks, such as avoiding hot water in a mug or preparing to open a bottle cap (\cref{fig:teaser} left). Hence, relying solely on the geometric information of objects is inadequate. Combining the semantic information of the object with the description of intent enables the generation of more natural and logical grasps.

Typical grasp representations in previous grasp generation methods exhibit constraints in embedding semantic information. For example, methods that depict grasps through robotic hand poses \cite{liu2020deep, zhu2021toward, xu2023unidexgrasp, wan2023unidexgrasp++, lu2023ugg, liu2024realdex, jin2024reasoning, tang2023graspgpt}, MANO model \cite{MANO:SIGGRAPHASIA:2017} based poses \cite{hasson2019learning, liu2021synthesizing, yang2022artiboost, christen2022d, turpin2022grasp, jiang2021hand, taheri2020grab, corona2020ganhand, taheri2022goal, wu2022saga, jian2023affordpose, zheng2023cams}, contact regions \cite{yang2022oakink, liu2023contactgen, grady2021contactopt, brahmbhatt2019contactgrasp, lakshmipathy2023contact, zhang2021manipnet, li2022contact2grasp, yang2024learning}, and implicit forms \cite{karunratanakul2020grasping, karunratanakul2021skeleton}. Attempts to generate dexterous grasps based on semantic cues, such as those by \cite{zhu2021toward, yang2022oakink, jian2023affordpose, xu2023unidexgrasp, liu2024realdex}, rely on coarse affordance vectors for conditional generation or directly use vision-language models \cite{radford2021learning, gemini} to filter sampled grasps. Nevertheless, integrating detailed semantic information or language descriptions into the grasp generation process remains challenging.

When humans plan a grasping posture, they initially determine the grasp's general \textit{orientation}, guided by the object's category and the semantics of the instruction. Subsequently, the specific \textit{manner} of grasping is decided, influenced by both the manipulation intent and the object's shape. Finally, the \textit{refinement} of the grasp pose is conducted, taking into account the object's detailed geometry and the hand-object contact state to ensure physical plausibility. Therefore, it is crucial to design a grasp representation that explicitly incorporates these three steps while implicitly embedding semantic information.

In this paper, we introduce a novel grasp generation method, termed \method, that incorporates semantic information into the generation process. This approach, inspired by human grasp planning, divides the grasp representation into three interrelated components: 1) \textit{orientation}, influenced by the intent and the object's function; 2) \textit{manner}, specifying the grasp taxonomy required for interaction; and 3) \textit{refinement}, detailing the hand pose adjustments necessary for physical plausibility.

Given that language is inherently discrete, we naturally extend this discreteness to other modalities to synchronize with the semantic space, similar to other multimodal studies \cite{jiang2023motiongpt, li2023videochat, yin2023shapegpt}. Furthermore, human grasp poses can be categorized into 33 discrete types according to the Grasp Taxonomy \cite{feix2015grasp}. By combining object shape and manipulation intent, a human grasp can be derived by adjusting these foundational types. Consequently, we employ a vector quantized variational autoencoder (VQ-VAE) \cite{van2017neural, razavi2019generating} to discretize the grasp components into tokens. This approach not only augments the alignment of grasp-language within the semantic space but also offers significant benefits: 1) it enhances grasp generation's controllability and interpretability through explicit token representation; 2) it markedly reduces the dimensionality of the grasp space, simplifying the learning process of this representation.
The VQ-VAE operates as an encoder-decoder structure, taking the object's point cloud as conditional input, whereby the encoder encodes the grasp into three tokens via a codebook lookup, and the decoder reconstructs these tokens back into the original grasp.

To align the discrete grasp representation of \method with semantic space more effectively, we leverage a multimodal large language model (MLLM). Inspired by the structure of LLaVA \cite{liu2023llava}, our MLLM takes the discretized grasp tokens, the object features obtained via PointBERT \cite{yu2022point}, and the language description as the input. Following many MLLM works \cite{liu2023llava, jiang2023motiongpt}, our model is trained in two training stages: 1) multimodal alignment, wherein the MLLM is trained to predict grasp tokens based on object features and language descriptions, thereby mapping these modalities in a unified space; and 2) instruction tuning, where the MLLM is fine-tuned to enhance grasp generation for more complex outputs.

Currently, well-aligned language-grasp datasets are scarce. The annotations in \cite{yang2022oakink, jian2023affordpose} cover only simple intentions. To train \method, we collect \dataset, extending existing hand-object interaction datasets in three ways: 1) Low-level annotations, identifying contact states, such as which fingers touch the object and which parts of the object are being grasped. These details are deduced from the positions of hand and object. 2) High-level annotations, encompass manipulation intent and grasp force, for instance, `tightly touch the bottle cap to unscrew it'. Utilizing low-level information, we generate these grasp-related descriptions with GPT-4 \cite{openai2023gpt}. 3) Conversational annotations: Employing GPT-4 and GPT-4v \cite{GPT4Vision23}, we construct grasp-language mixed conversations based on dataset images or rendered visuals. These dialogues include both low and high-level information and infer grasp details from the images. We train \method using \dataset. Experiments demonstrate that our method well generate the corresponding grasp pose across multiple metrics. We also verify our method's potential value in AR/VR and embodied robotics applications.

In summary, our contributions are threefold: 1) We propose \method, an innovative grasp generation method integrating semantic information. 2) We introduce a novel grasp discrete representation, efficiently and effectively expressing grasp postures and supporting the task of grasp description through language alignment. 3) We compile \dataset dataset. To the best of our knowledge, it is the first dataset of semantic grasps that encompass low-level, high-level, and conversational annotations.
\section{Related Works}\label{sec:related_works}
\cheading{Grasp Generation}
Grasp generation remains a fundamental task with wide applications in robotics \cite{bohg2013data, newbury2023deep}. Recently, the generation of human-like grasps has attracted increasing attention. Unlike the 6DoF (degrees of freedom) parallel-jaw grippers commonly used in robotics \cite{eppner2021acronym, mousavian20196, fang2020graspnet, fang2023anygrasp, tang2023graspgpt, jin2024reasoning}, the higher freedom in human fingers significantly complicates grasp generation. ObMan \cite{hasson2019learning} leverages GraspIt! \cite{miller2004graspit} for synthesizing grasps, while \cite{liu2021synthesizing} optimize grasp based on force closure. Methods such as \cite{christen2022d, xu2023unidexgrasp, wan2023unidexgrasp++} validate their techniques in physical simulations. Data-driven approaches \cite{hasson2019learning, liu2021synthesizing, christen2022d, turpin2022grasp, jiang2021hand, taheri2020grab, corona2020ganhand, taheri2022goal, wu2022saga, jian2023affordpose, zheng2023cams, liu2024taco, yang2024learning} employ end-to-end generative models like cVAE \cite{sohn2015learning} or GAN \cite{goodfellow2014generative}. Given the high degrees of freedom in the MANO model \cite{MANO:SIGGRAPHASIA:2017}, most of these works implement post-processing to enhance contact consistency and physical plausibility. However, these methods solely emphasize geometric information of objects or incorporate basic intent features \cite{yang2022oakink,jian2023affordpose}. In contrast, our \method integrates semantic aspects of grasping, facilitating the direct generation of grasp postures that correspond with language descriptions.

\cheading{Hand-Object Interaction Datasets}
Understanding hand-object interaction is pivotal in AR/VR, animation, and embodied AI. Existing datasets primarily concentrate on hand-object pose estimation or reconstruction. These datasets are either synthesized through rendering techniques \cite{hasson2019learning, gao2022dart, li2023chord, corona2020ganhand} or compiled by annotating real-world data \cite{garcia2018first, taheri2020grab, brahmbhatt2020contactpose, hampali2020honnotate, hampali2022keypoint, sener2022assembly101, chao2021dexycb, qin2022dexmv, yang2022oakink, liu2022hoi4d, kwon2021h2o, xie2023hmdo, fan2023arctic, zhu2023contactart, liu2024taco, Li_Yang_Lin_Xu_Zhan_Zhao_Zhu_Kang_Wu_Lu_2024, kim2024parahome, zhan2024oakink2}. Some studies delve into the semantics of grasping. For instance, \cite{zhu2021toward} introduces `touch codes' to depict contact states between fingers and object parts. \cite{yang2022oakink} offers annotations of grasping intent and object segmentation based on affordances. \cite{jian2023affordpose} segments object point clouds according to grasping semantics. Nonetheless, these datasets typically provide only basic semantic categorizations. Our \dataset, in contrast, delivers comprehensive annotations of hand-object interaction, encompassing detailed low-level contact state information, high-level grasp-related descriptions, and conversational annotations.

\cheading{Multimodel Large Language Models}
The development and application of Large Language Models (LLMs) have surged in recent years. Leading commercial models like GPT-4 \cite{openai2023gpt} and open-source counterparts such as Llama \cite{touvron2023llama} and Vicuna \cite{zheng2023judging} demonstrate exceptional language understanding and generation capabilities. There's an increasing trend of integrating LLMs into multimodal tasks, including image \cite{liu2023llava, zhu2023minigpt, huang2023language}, video \cite{zhang2023video, li2023videochat}, 3D \cite{xu2023pointllm, huang2023embodied, yin2023shapegpt}, and human posture and motion tasks \cite{jiang2023motiongpt, feng2023posegpt}. For example, \cite{liu2023llava} uses a vision encoder to extract image features and aligns these with the language space using projection layers. \cite{jiang2023motiongpt} interprets motion sequences as a series of tokens, and fine-tunes the T5 model \cite{raffel2020exploring} with LoRA \cite{hu2021lora} to facilitate various tasks like motion generation and captioning. Our work focuses on fine-tuning a model based on Vicuna \cite{zheng2023judging} for tasks related to grasp generation, harnessing LLMs' power to interpret and generate complex hand-object interactions.
\section{Method}
\label{sec:method}

\subsection{Overview}
\label{sec:overview}

\begin{figure}[tb]
    \centering
    \includegraphics[width=\linewidth]{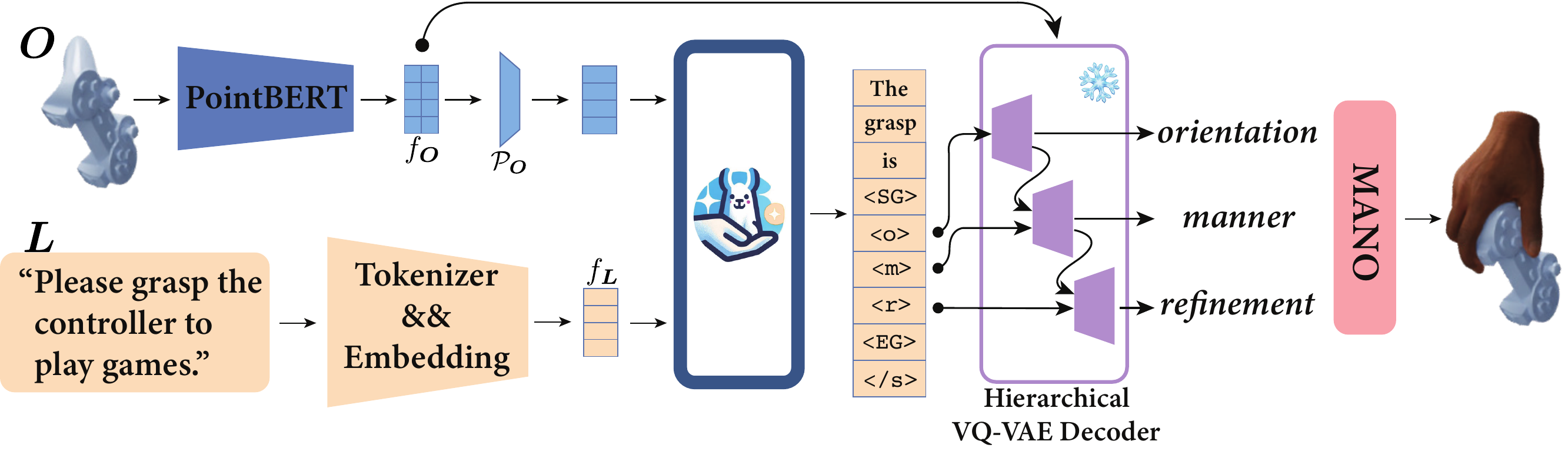}
    \caption{Our \method~pipeline. The grasp-aware language model outputs both the grasp tokens and the language conversations.}
    \label{fig:pipeline}
\end{figure}

Given a specific object $\obj$, represented as a point cloud, our goal is to align the human grasp $\grasp$ with the associated language description $\lang$, facilitating the task of semantic grasp generation. To this end, we introduce a novel grasp generation methodology, termed \method, which fundamentally comprises two principal components: grasp discretization (\cref{sec:grasp_discretization}) and a grasp-aware language model (\cref{sec:grasp_aware_language_model}). The initial phase involves the tokenization of the grasp into three interrelated tokens by training a VQ-VAE \cite{van2017neural, razavi2019generating}. As shown in \cref{fig:teaser} right, after the tokenizer is trained, the VQ-VAE is frozen, enabling the projection of \grasptoken tokens from the codebook onto grasp configurations. Subsequently, as illustrated in \cref{fig:pipeline}, the grasp-aware language model is designed to reconcile the discrete grasp representations with the linguistic domain, trained specifically to generate corresponding \grasptoken tokens\del{ and linguistic captions}. These resultant \grasptoken tokens can then be reverted to the original grasp pose through the VQ-VAE decoder. The training of the grasp-aware language model is conducted utilizing our dataset, \dataset, which builds upon existing dataset of hand-object interactions \cite{yang2022oakink}, augmented through automated expansions (\cref{sec:collect_dataset}).

\subsection{Grasp Discretization}
\label{sec:grasp_discretization}

Consistent with prior studies \cite{karunratanakul2020grasping, jiang2021hand}, we define the grasp $\grasp = (\mbold{T}, \mbold{\theta}, \mbold{\beta})$ within the canonical space of the object. Here, $\mbold{T} \in \real^{4\times 4}$ represents the homogeneous transformation matrix, indicating the global rotation and translation of the hand relative to the object's central coordinate system. The parameters $\mbold{\theta} \in \real^{15\times 3}$ and $\mbold{\beta} \in \real^{10}$ denote the local hand pose and shape parameters, respectively. The hand vertices $\hand \in \real^{778 \times 3}$ are computed using a differentiable layer, specifically the MANO $\mathcal{M}$ model \cite{MANO:SIGGRAPHASIA:2017}, where $\hand = \mathcal{M}(\grasp) = \mathcal{M}(\mbold{T}, \mbold{\theta}, \mbold{\beta})$.

\begin{figure}[t]
    \centering
    \includegraphics[width=0.95\linewidth]{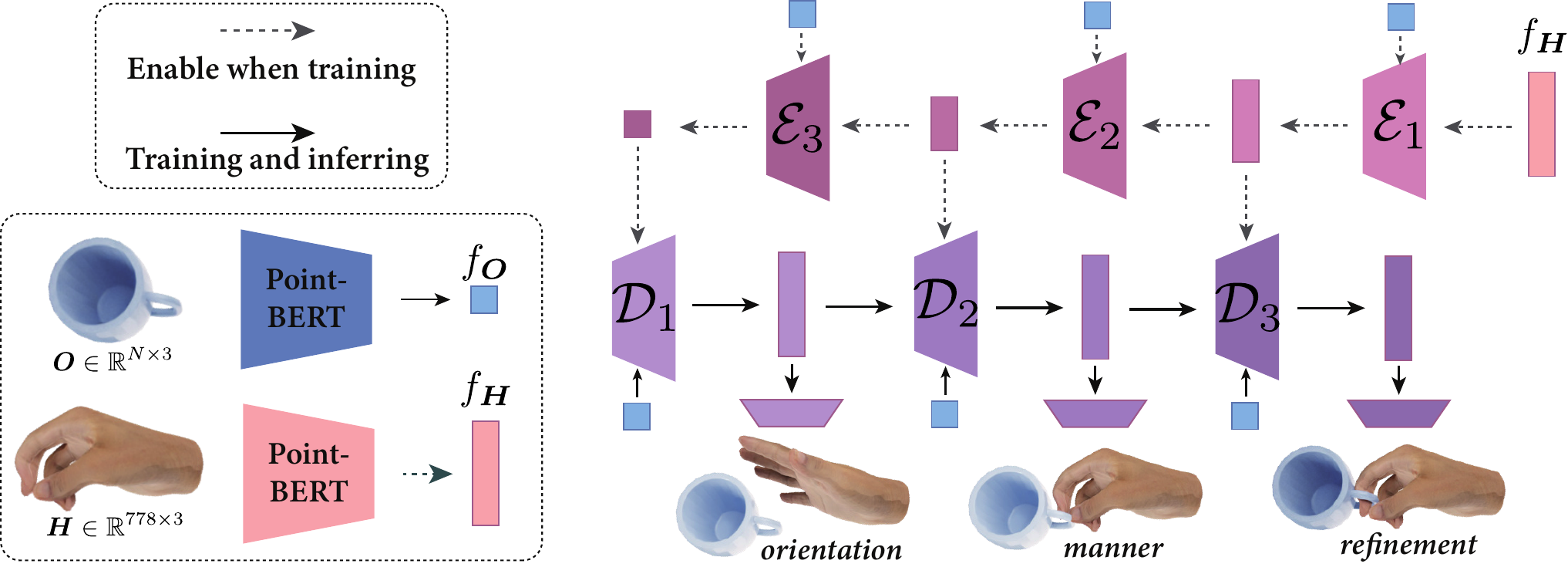}
    \caption{Our grasp discretization process is based on hierarchical VQ-VAE.}
    \label{fig:grasp_tokenizer}
\end{figure}

In this work, to more effectively illustrate the human grasp process and align it with the semantic space, we discretize the grasp $\grasp$ into three components \codebracket{$\tokenO, \tokenM, \tokenR$}, representing the \textit{orientation}, \textit{manner}, and \textit{refinement} token, respectively, where $\tokenO, \tokenM, \tokenR \in \mathbb{N}$. We employ a hierarchical VQ-VAE \cite{razavi2019generating}, encompassing the trainable codebooks $\mathcal{B}_i$, encoders $\mathcal{E}_i$, and decoders $\mathcal{D}_i$, where $i \in \{1, 2, 3\}$, to quantize the grasp vector into meaningful integers and subsequently reconstruct the original grasp vector from the quantized tokens (\cref{fig:grasp_tokenizer}).

The encoders progressively map the hand's representation into the latent space, capturing grasp information from low to high levels. This structured approach enables the simulation of the grasping process through conditional probabilities: 1) The hand's global information $\mbold{T}$ is captured with the \textit{orientation} token \codebracket{$\tokenO$}, where $\hat{\mbold{T}} = \mathcal{D}_1(\tokenO, \obj)$. 2) The local hand pose $\mbold{\theta}, \mbold{\beta}$ is encapsulated by the \textit{manner} token \codebracket{$\tokenM$}, conditioned on \codebracket{$\tokenO$}, with $\hat{\mbold{\theta}}, \hat{\mbold{\beta}} = \mathcal{D}_2(\tokenO, \tokenM, \obj)$. 3) For the fine-tuning process, the delta parameters $\Delta\mbold{T}, \Delta\mbold{\theta}, \Delta\mbold{\beta}$ are represented by the \textit{refinement} token \codebracket{$\tokenR$}, conditioned on \codebracket{$\tokenO, \tokenM$}, where $\Delta\hat{\mbold{T}}, \Delta\hat{\mbold{\theta}}, \Delta\hat{\mbold{\beta}} = \mathcal{D}_3(\tokenO, \tokenM, \tokenR, \obj)$. The final grasp is reconstructed as $\hat{\grasp} = (\Delta\hat{\mbold{T}} \cdot \hat{\mbold{T}}, \Delta\hat{\mbold{\theta}} + \hat{\mbold{\theta}}, \Delta\hat{\mbold{\beta}} + \hat{\mbold{\beta}})$. The hat symbol $\hat{\cdot}$ denotes the reconstructed values.

Specifically, the codebook $\mathcal{B} = \{\mbold{b}_k\}_{k=1}^{K}$, with each $\mbold{b}_k \in \real^{d_\mathcal{B}}$, where $d_\mathcal{B}$ represents the dimension of the VQ-VAE latent space and $K$ the number of codebook entries. For each input vector $\mbold{z}$, the encoder $\mathcal{E}$ maps it to the latent space using the mapping network $\mathcal{N}_{\mathcal{E}}$ and finds the nearest codebook entry $\mbold{b}_{\tokenZ}$: 
$\tokenZ = \mathcal{E}(\mbold{z}) = {\mathrm{argmin}}_{k} \|\mathcal{N}_{\mathcal{E}}(\mbold{z}) - \mbold{b}_k\|_2$, where $\tokenZ \in \{1,2,\ldots, K\}$.
The decoder $\mathcal{D}$ then reconstructs the original vector $\mbold{z}$ with the mapping network $\mathcal{N}_{\mathcal{D}}$:
$\hat{\mbold{z}} = \mathcal{D}(\tokenZ) = \mathcal{N}_{\mathcal{D}}(\mbold{b}_{\tokenZ})$. The VQ-VAE is trained to minimize the reconstruction loss $\mathcal{L}_{\text{rec}}$, the embedding loss $\mathcal{L}_{\text{emb}}$, and the commitment loss $\mathcal{L}_{\text{com}}$:
\begin{equation}
    \mathcal{L}_{\text{rec}} = \|\mbold{\hand} - \hat{\hand}\|_2^2 = \|\mbold{\hand} - \mathcal{M}(\hat{\grasp})\|_2^2
\end{equation}

\begin{equation}
    \mathcal{L}_{\text{emb}} + \mathcal{L}_{\text{com}} = \|\mathrm{sg}[\mathcal{N}_{\mathcal{E}}(\mbold{z})] - \mbold{b}_{\tokenZ}\|_2^2 + \|\mathcal{N}_{\mathcal{E}}(\mbold{z}) - \mathrm{sg}[\mbold{b}_{\tokenZ}]\|_2^2
\end{equation}
where $\mathrm{sg}[\cdot]$ denotes the stop-gradient operation.

\subsection{Grasp Aware Language Model}
\label{sec:grasp_aware_language_model}

\begin{figure}[t!]
    \centering
    \includegraphics[width=\linewidth]{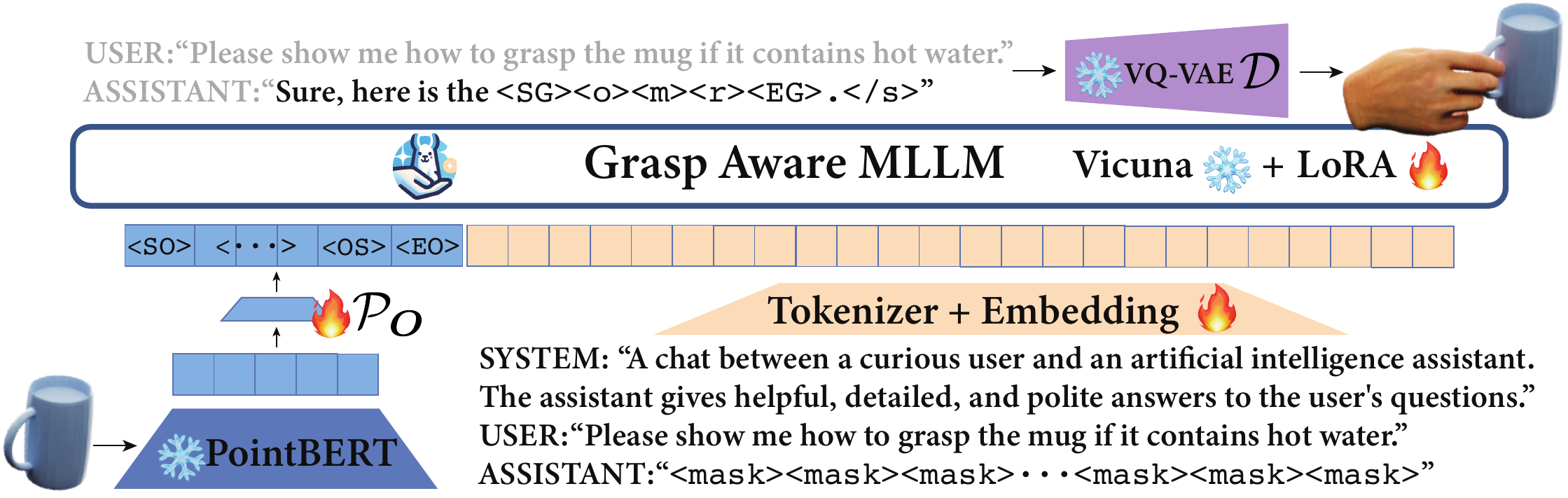}
     \caption{Grasp Aware Language Model.}
     \label{fig:MLLM}
\end{figure}

Building upon the grasp discrete representation, we design a grasp-aware language model aimed at facilitating semantic grasp generation tasks. As depicted in \cref{fig:MLLM}, our model is trained to align three distinct modalities: the human grasp $\grasp$, object points $\obj$, and the language description $\lang$.

\cheading{Grasp Modal}
After we train the VQ-VAE, this module is frozen. We take the encoders $\mathcal{E}$ as the grasp tokenizer that transfers the grasp $\grasp$ into the \grasptoken token which contains three components: \codebracket{$\tokenO, \tokenM, \tokenR$}. To distinct the grasp token from the language, we add special tokens $\texttt{\codebracket{SG}}$ and $\texttt{\codebracket{EG}}$ to the start and end of the grasp, respectively. The grasp token, as generated by the MLLM model, can be converted back to the human grasp utilizing the VQ-VAE decoders $\mathcal{D}$.

\cheading{Object Modal}
We employ PointBERT \cite{yu2022point} to extract the object features $f_{\obj}$ from the point cloud $\obj$, where $\obj \in \real^{N \times 3}$ and $f_{\obj} \in \real^{M \times d_{\obj}}$. Here, $N$ denotes the number of points, $M$ the count of point features, and $d_{\obj}$ the dimension of the object feature space. Notably, PointBERT requires the normalization of object sizes, a factor crucial for grasp generation. Thus, we incorporate the object size as a distinct token $\texttt{\codebracket{OS}}$ within the MLLM inputs. Subsequently, object features are projected into a unified space alongside the grasp token via a linear projection layer $\mathcal{P}_{\obj}$. Similar to the grasp token, special tokens $\texttt{\codebracket{SO}}$ and $\texttt{\codebracket{EO}}$ are affixed to the object feature sequence, delineating its start and end. The projected object features are treated as the system message of the GPT's style conversational prompt.

\cheading{Language Modal}
Our language model undergoes fine-tuning based on the Vicuna-7B checkpoint \cite{zheng2023judging}, with a tailored prompt designed to direct the completion of specified tasks (detailed further in \supp). Textual content is tokenized into 32K word pieces employing the SentencePiece methodology \cite{kudo2018sentencepiece}.

\cheading{Training}
Our model architecture, akin to LLaVA \cite{liu2023llava}, aligns all modalities through projection or embedding layers into a unified semantic space, yielding $X = \{\mbold{x}^i\}^{T}_{i=1}\subset \real^T_{d_\mathcal{L}}$, where $d_\mathcal{L}$ signifies the semantic space dimensionality. An autoregressive language model is then trained to optimize the likelihood of the next tokens, predicated on preceding ones $p(\hat{X} | X) = \prod_i p(\hat{x}^i | \hat{x}^{<i}, x)$, terminating with the end-of-sentence token $\texttt{</s>}$. The primary objective is to diminish the negative log-likelihood loss $\mathcal{L}_{\text{NLL}}$:
\begin{equation}
\mathcal{L}_{\text{NLL}} = -\log p(\hat{X} | X) = -\sum_i \log p(\hat{x}^i | \hat{x}^{<i}, x)
\end{equation}
Utilizing LoRA \cite{hu2021lora}, we finetune the MLLM model, adjusting approximately 0.4B parameters within the linear layers. This process unfolds in two phases: 1) multimodal alignment, wherein the MLLM is trained to predict the grasp tokens based on object features and language descriptions, thereby consolidating these modalities within a unified framework. During this phase, the object feature projection layer $\mathcal{P}_{\obj}$ is updated. The embedding layer of the MLLM is also refined to accommodate the newly introduced special tokens; and 2) instruction tuning, throughout which the MLLM undergoes further refinement for grasp generation task combined with language outputs, with the projection layer being frozen to ensure training stability.

\subsection{\dataset Dataset}
\label{sec:collect_dataset}

Currently, there exists no dataset with well-aligned grasp language annotations that would enable us to train \method for supporting grasp generation tasks and the downstream applications. Considering the prohibitive costs and labor-intensive nature of manual semantic annotation, we design an automatic annotation methodology based on GPT-4 to augment existing hand-object interaction datasets. Our dataset, \dataset, encompasses low-level, high-level, and conversational annotations.

\cheading{Low-level Annotations}
Low-level annotations refer to the contact relationships between each finger and various parts of the object. According to the Grasp Taxonomy \cite{feix2015grasp}, we can deduce the grasp type and intent from these low-level annotations. For instance, if the thumb and index finger are in contact with a screw, it is inferred that the grasp type is a `pinch' and the intent is to `screw/unscrew'. The OakInk dataset provides annotations for objects' CAD models, hand vertices, and object part segmentation. Utilizing this information, we calculate the contact states when the distance between hand vertices and the object's part segmentation points is less than a threshold of $3mm$.

\cheading{High-level Annotations}
High-level intent is annotated from two perspectives: 1) based on low-level contact information. Given this information (i.e., the finger and object part contact), we employ GPT-4 to infer the grasping intent. For example, if all fingers are grasping the handle of a mug, GPT-4 can deduce that it is a firm grasp with possible intents such as `make a toast' or `avoid hot beverage'. 2) Based on images or rendered views. Since the OakInk dataset includes a subset of real captured images (i.e., OakInk-image), we manually select representative frames that are clear, unobstructed, and with explicit intent. We leverage GPT-4v, a commercial visual-language model, to infer high-level information such as manipulation intent and grasp force. For grasps in OakInk without matching images (i.e., OakInk-shape), we render the image using the Blender renderer with realistic hand textures \cite{li2022nimble}. The details of the prompt are elaborated in \supp.

\cheading{Conversational Annotations}
With the aforementioned low-level and high-level annotations, we construct conversations using the GPT-4 model. We ask GPT-4 to generate various conversational templates from different perspectives, including detailed hand-object contact information, manipulation intent, grasp force, and type. These dialogues must be consistent with the grasp and ensure logical plausibility. The prompts that guide the GPT-4 model are detailed in \supp.

Considering the hallucination problem of GPT-4, to ensure the quality of our \dataset, we manually review these annotations to filter out intents that defy common sense and conversations that lack logical coherence. Statistically, our dataset includes approximately 1.8k object models from OakInk, about 50,000 hand-object grasp pairs. For each pair, we offer on average 5 detailed captions and conversational annotations.
\section{Metrics and Experiments}
\label{sec:experiments}
Our methodology, \method, incorporates two principal components: the grasp discrete representation and the grasp-aware language model. We assess the reconstruction accuracy of VQ-VAE to demonstrate the validity of our grasp discretization approach. Additionally, we evaluate the performance of grasp generation by our MLLM. Comparative and ablation studies underscore the effectiveness of our methodology.

\subsection{Metrics}

\cheading{Aspect of Physical Plausibility}
To evaluate the physical plausibility of the predicted grasp pose $\hat{\grasp}$, we employ several metrics: 1) Mean Per-Vertex Position Error (MPVPE, in $mm$) calculates the average $L2$ distance per vertex between the predicted hand mesh $\hat{\hand}$ and the ground truth $\hand$, when available. 2) Penetration Depth (PD, in $cm$) measures the maximum penetration depth of hand vertices into the object, indicating surface penetration. 3) Solid Intersection Volume (SIV, in $cm^3$) quantifies the volumetric intersection by voxelizing the object mesh and calculating the volume within the hand surface. 4) Simulation Displacement (SD, in $cm$) tests grasp stability in PyBullet \cite{coumanspybullet}, measuring the object's center displacement under steady hand conditions and gravity \cite{hasson2019learning}. These metrics gauge both the quality of grasp generation and the accuracy of our grasp discrete representation.

\cheading{Aspect of Semantic Consistency}
Semantic consistency is evaluated by examining the quality of grasp generation: 1) GPT-4 assisted evaluation. For generated grasps $\hat{\grasp}$, we first render the hand-object interaction following the same pipeline as in \cref{sec:collect_dataset}. Then, we use GPT-4v to score the semantic consistency of the grasp images based on input captions. Scores range from 0 to 100, with higher scores indicating better consistency. The prompts used in GPT-4 assisted evaluation are listed in \supp. 2) P-FID calculates the Fréchet Inception Distance between the point clouds of the $\hat{\hand}$ and $\hand$, using the pre-trained feature extractor from \cite{nichol2022point}. 3)\del{ Traditional NLP metrics for caption generation include BLEU-4 \cite{papineni2002bleu} and CIDEr \cite{vedantam2015cider}, measuring n-gram overlap and similarity based on cosine similarity, respectively. 4)} Perceptual Score (PS) assesses the naturalness of grasps and semantic consistency, with 5 volunteers rating the generated grasps on a 5-point Likert scale. The final score is the mean Likert score.

\subsection{Implementation Details}
The VQ-VAE's codebook $\mathcal{B}$ consists of $K = 512$ entries, each dimensioned at $d_\mathcal{B} = 256$. We employ PointBERT \cite{yu2022point} as the point cloud feature extractor in the VQ-VAE encoder $\mathcal{E}$ for both hand vertices $\hand$ and object vertices $\obj$. Similar to \cite{xu2023pointllm}, PointBERT is pretrained using the ULIP-2 method \cite{xue2023ulip} for enhanced geometry-language alignment. The predicted rotation from the VQ-VAE decoders $\mathcal{D}$ uses a 6D representation \cite{zhou2019continuity}, subsequently converted to the axis-angle representation for further computation.

For the MLLM, we utilize Llama structure \cite{touvron2023llama} as the model backbone, finetuning based on the Vicuna-7B checkpoint \cite{zheng2023judging}. To extract object feature $f_{\obj}$, we reuse PointBERT from the VQ-VAE and freeze its parameters. $f_{\obj}$ includes $M = 513$ embeddings, each of dimension $d_{\obj} = 384$. The object projection layer $\mathcal{P}_{\obj}$ projects $f_{\obj}$ to the language space dimension of $4096$. We configure the LoRA module \cite{hu2021lora} with a rank setting of $r = 64$, resulting in approximately 6\% parameters being finetuned. The learning rate is set to 5e-4 and 3e-5 for the multimodal alignment stage and instruction tuning stage, respectively, with a cosine annealing learning rate scheduler for training stability. The batch size is 128, and the MLLM is trained over 20 epochs on 4 A100 GPUs with 80GB of memory each.

\subsection{Comparisons}

\begin{table}[b]
    \caption{Our discrete VQ-VAE grasp representation compared with SOTA methods.}
    \label{tab:vqvae_cmp}    
    \vspace{-0.5em}
    \centering
    \setlength{\tabcolsep}{5pt}{
        {\scriptsize
        \begin{tabular}{@{}l|c|c|c|c|c@{}}
            \toprule        
                            & MPVPE $\downarrow$ & PD $\downarrow$ & SIV $\downarrow$ & SD \textit{mean.} $\downarrow$ & SD \textit{std.} $\downarrow$ \\
            \midrule
            \dataset  dataset                                   &   -        &   0.11    &   0.62    &   0.94    &   1.62\\
            \midrule
            GrabNet \cite{taheri2020grab} \textit{w/o refineNet}    &   18.14    &   0.76    &   5.42    &   1.75    &   2.61\\
            GrabNet \cite{taheri2020grab}                           &   27.49    &   0.54    &   3.45    &   1.77    &   2.36\\
            GrabNet \cite{taheri2020grab} \textit{w/ TTA}           &   27.16    &   0.49    &   2.16    &   \textbf{1.35}    &  1.56\\
            \midrule
            Jiang \etal \cite{jiang2021hand}  \textit{w/o TTA}      &   33.68    &   0.72    &   5.81    &   1.53    &   1.77\\
            Jiang \etal \cite{jiang2021hand}  \textit{w/ TTA}       &   33.84    &   0.58    &   2.78    &   1.36    &   \textbf{1.55}\\
            \midrule
            Ours \textit{w/o finetune token}                          &   20.36    &   0.48    &   3.00    &   1.95    &   2.11\\
            Ours                                                    &   \textbf{14.97}    &   0.46    &   2.72    &   2.14    &   2.37\\
            Ours \textit{w/ TTA}                                    &   23.61    &   \textbf{0.37}    &   \textbf{1.27}    &   1.90    &   2.12\\
            \bottomrule
        \end{tabular}
        }
    }
\end{table}

\cheading{Discrete VQ-VAE Grasp Representation} 
Given the discretization of grasps, three primary concerns arise: 1) Does discretization compromise reconstruction accuracy? 2) Does it affect the physical plausibility of the interaction?, and 3) Does it have the capability to embed semantic information?

To answer the first two questions, we compare our method with two state-of-the-art methods, GrabNet \cite{taheri2020grab} and Jiang \etal \cite{jiang2021hand}, on a reconstruction task. Both methods are based on cVAEs \cite{sohn2015learning} for grasp generation. GrabNet employs RefineNet to refine the interaction in an end-to-end iterative manner, whereas Jiang \etal utilize test time adaptation (TTA) to optimize hand-object contact. Our method leverages a \textit{refinement} token to adjust the hand pose in an end-to-end manner. Compared to configurations without the \textit{refinement} token, our approach with the \textit{refinement} token exhibits superior performance, achieving a 26\% improvement in MPVPE and a 9\% improvement in SIV. Considering the TTA, an optimization-based approach, can precisely improve hand-object interaction, we also report our method's performance with TTA in \cref{tab:vqvae_cmp} for a fair comparison, which attains current SOTA results in PD and SIV. The results demonstrate that our discrete grasp representation method can accurately depict hand poses and specifically ensure the physical plausibility of interactions. Compared to previous SOTA methods, our method is competitive and exhibits advantages in certain metrics.

\begin{figure}[tb]
    \centering
    \begin{subfigure}{0.35\linewidth}
        \includegraphics[width=\linewidth]{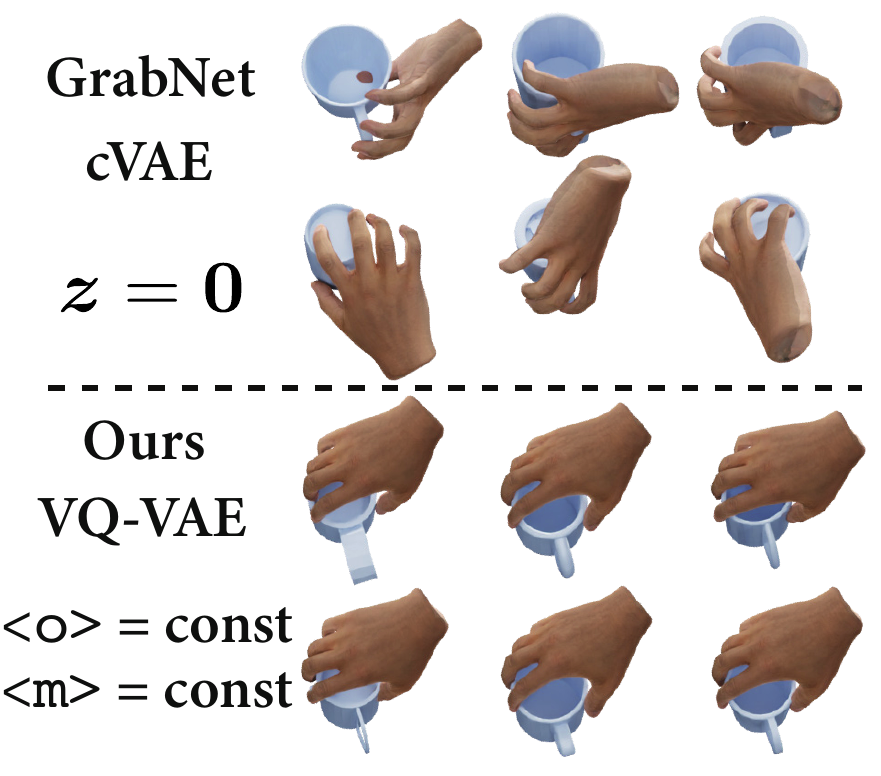}
      \caption{Controllable generation compared with cVAE.}
      \label{fig:sem_cmp}
    \end{subfigure}
    \hfill
    \begin{subfigure}{0.62\linewidth}
      \includegraphics[width=\linewidth]{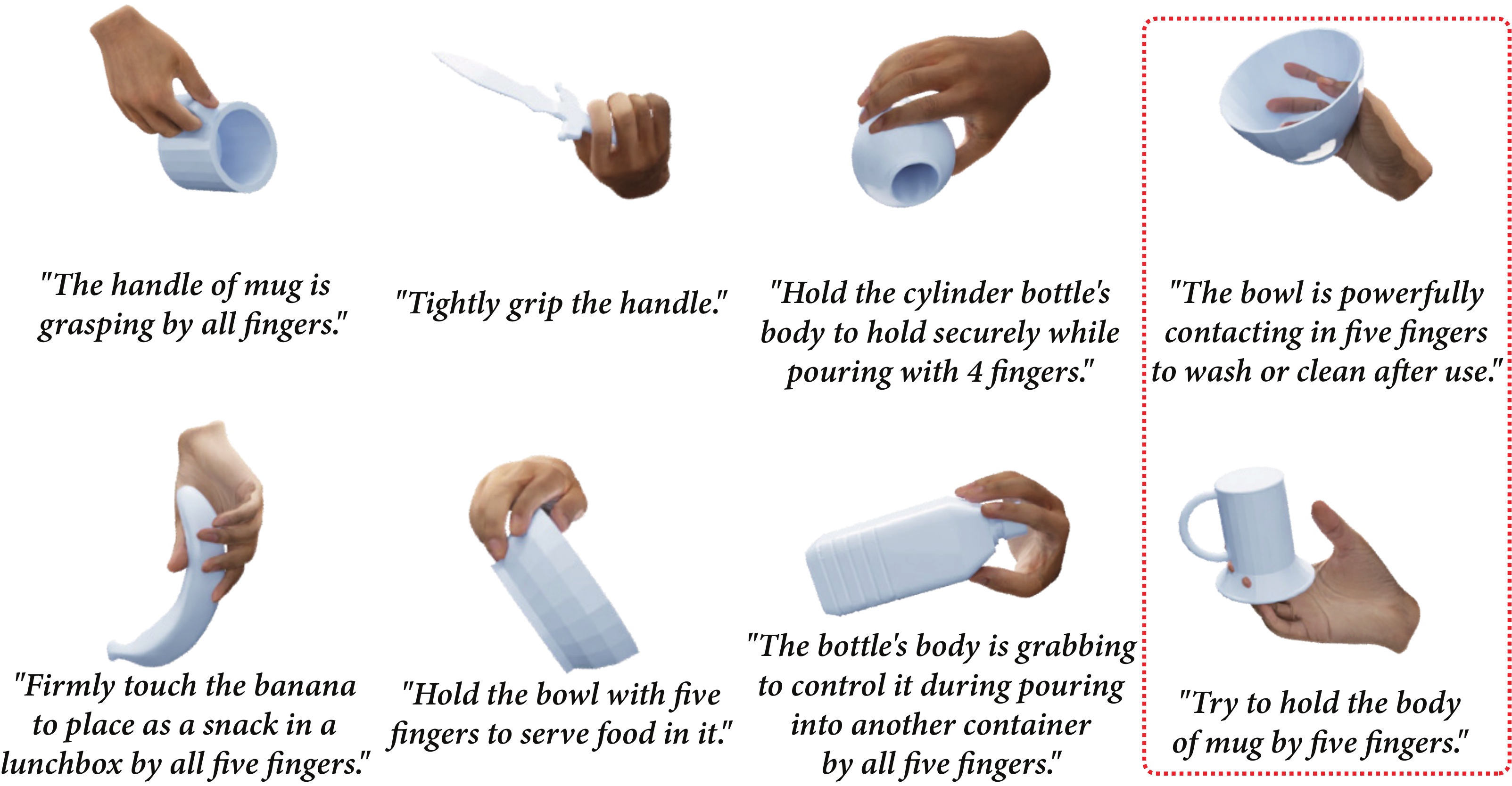}
      \caption{MLLM's qualitative results. We only visualize grasp and ignore the text outputs for better visualization. The red box indicates the failure cases.}
      \label{fig:qualitative}
    \end{subfigure}
    \caption{Qualitative results of our \method.}
\end{figure}

To explore whether our discrete representation method can encode semantic information, we conduct a controllable generation task. For GrabNet, the cVAE-based method, we directly fix the sampling vector $\mbold{z} = \mbold{0}$, enforcing the model to generate similar grasps. For our method, we assign the same const value to the \codebracket{$\tokenO, \tokenM$} tokens. As depicted in \cref{fig:sem_cmp}, take the specific category of mugs as an example, although the shape of the mug varies, our generated grasps maintain consistency in orientation and manner, showing the semantic consistency of our method. In contrast, we find that GrabNet's prediction results are not interpretable.

\begin{table}[tb]
    \caption{Quantitative results of our MLLM based grasp generation method.}
    \label{tab:mllm_grasp_gen}
    \vspace{-0.5em}
    \centering
    \setlength{\tabcolsep}{5pt}{
        {\scriptsize
        \begin{tabular}{@{}l|c|c|c|c|c|c|c@{}}
            \toprule        
                            & P-FID $\downarrow$ & PD $\downarrow$ & SIV $\downarrow$ & SD \textit{mean.} $\downarrow$ & SD \textit{std.} $\downarrow$ &GPT-4 $\uparrow$ & PS $\uparrow$ \\
            \midrule
            \dataset                                   &   -        &   0.11    &   0.62    &   0.94    &   1.62 &  82.3 & 4.7\\
            \midrule
            BERT \cite{devlin2018bert} based  &   3.32    &   0.49    &    4.60     &   2.17 & \textbf{2.26} & 47.3    &   3.7\\
            \method     &   \textbf{2.28}    &   \textbf{0.48}    &   \textbf{4.24}    & \textbf{2.00} & 2.33 &  \textbf{74.5}   &   \textbf{4.6}\\
            \bottomrule
        \end{tabular}
        }
    }
\end{table}

\cheading{Language Guided Grasp Generation}
We need to validate that MLLM can control grasp generation $\grasp$ based on textual input $\lang$. To the best of our knowledge, there are no existing works directly comparable to ours. Therefore, leveraging our discrete representation, we construct a straightforward baseline that treats this task as a classification problem. We finetune the official BERT model \cite{devlin2018bert} to embed the language description in conjunction with the object feature. Subsequently, we deploy three distinct classification heads to predict the \codebracket{$\tokenO, \tokenM, \tokenR$} tokens. We train this modified BERT with our \dataset following the same settings in \method. These predicted tokens are then decoded into the final grasp pose as in our \method. The outcomes of this experimental setup are documented in \cref{tab:mllm_grasp_gen}. From the results, we observe that our MLLM outperforms the baseline in both the physical plausibility and semantic consistency metrics. Showing that simply treating the task as a classification problem is not sufficient to generate grasp pose that aligns with the language description. On one hand, benefitted from the pretrained LLM, our \method can well understand the language instructions. On the other hand, the discrete representation is interpretable, making the MLLM more controllable. We demonstrate the qualitative results of our \method in \cref{fig:qualitative}.

\subsection{Ablation Studies}
\begin{table}[t]
    \begin{center}
        \begin{minipage}[!htp]{0.488\linewidth}
        \begin{center}
            \caption{Representation ablation.}
            \label{table:representation_ablation}
            \vspace{-1.4em}
            \resizebox{1.0\linewidth}{!}{
            \makeatletter\def\@captype{table}\makeatother
            \setlength{\tabcolsep}{2pt}{
                \begin{tabular}{@{}l|c|c|c|c|c@{}}
                    \toprule        
                    \multirow{2}{*}{} & \multirow{2}{*}{MPVPE $\downarrow$} & \multirow{2}{*}{PD $\downarrow$} & \multirow{2}{*}{SIV $\downarrow$} & \multicolumn{2}{c}{SD} \\
                    \hhline{~~~~--}
                    &  &  &  & \textit{mean.} $\downarrow$  & \textit{std.} $\downarrow$  \\
                    \midrule
                    One token                                        &   29.95      &  0.66    &   5.14   &   1.90    &   2.29\\
                    \del{Two token}\codebracket{$\tokenO, \tokenM$}    &   25.73    &   0.58    &   4.32    &   2.14    &   2.50\\
                    \del{Multi-refine}\codebracket{$\tokenO, \tokenM, \tokenR \times 2$}  &   15.37    &   0.50    &   2.98    &   2.28    &   2.57\\
                    \del{Multi-refine}\codebracket{$\tokenO, \tokenM, \tokenR \times 3$}           &   15.90    &   0.52    &   3.37    &   1.98    &   2.18\\
                    Single VQ &   28.02    &   0.68   &   5.31    &   1.81    &   2.00\\
                    \textit{w/o} semantic &   21.94    &   0.60    &   4.59    &   1.84    &   1.91\\
                    \bottomrule
                \end{tabular}
            }
        }
        \end{center}
        \end{minipage}
        \;
        \begin{minipage}[!htp]{0.488\linewidth}
        \begin{center}
            \caption{VQ-VAE settings.} 
            \label{table:vq_ablation}
            \vspace{-1.4em}
            \resizebox{1.0\linewidth}{!} {
            \makeatletter\def\@captype{table}\makeatother
            \setlength{\tabcolsep}{2pt}{
                \begin{tabular}{@{}l|c|c|c|c|c@{}}
                    \toprule        
                    \multirow{2}{*}{} & \multirow{2}{*}{MPVPE $\downarrow$} & \multirow{2}{*}{PD $\downarrow$} & \multirow{2}{*}{SIV $\downarrow$} & \multicolumn{2}{c}{SD} \\
                    \hhline{~~~~--}
                    &  &  &  & \textit{mean.} $\downarrow$  & \textit{std.} $\downarrow$  \\
                    \midrule
                    $\mathcal{B}$ entries $K = 256$         &   95.52        &   1.59    &   51.83    &   1.42    &   1.29\\
                    $\mathcal{B}$ entries $K = 1024$         &   39.82        &   0.80    &   5.30    &   3.10    &   3.40\\
                    $\mathcal{B}$ dim. $d_\mathcal{B} = 512$    &   46.32    &   1.11    &   10.86    &   2.00    &   3.09\\
                    $\mathcal{B}$ dim. $d_\mathcal{B} = 128$  &   38.67    &   0.80    &   6.42    &   2.28    &   2.78\\
                    \midrule
                    Vanilla &   48.48    &   1.09    &   9.42    &   2.24    &   2.87\\
                    EMA+Reset \cite{razavi2019generating}  &   24.67    &   0.73    &   5.74    &   1.81   &   2.00\\
                    \bottomrule
                \end{tabular}
            }
            }
        \end{center}

    \end{minipage}
    \end{center}\vspace{-2.5em}
\end{table}
\cheading{Ablation on Discrete Representation}
We conduct ablation studies to examine the design of our tokenization approach. As outlined in \cref{sec:grasp_discretization}, we utilize three tokens—\textit{orientation}, \textit{manner}, and \textit{refinement}—to represent $\grasptoken$ as \codebracket{$\tokenO, \tokenM, \tokenR$}. In our evaluation, detailed in \cref{table:representation_ablation}, we explore different token configurations: 1) Single Token: Compressing $\grasp$ into a single codebook significantly degrades reconstruction accuracy. 2) Two Tokens: This setup differs from the \textit{w/o refinement} token setting in \cref{tab:vqvae_cmp}, as here we train the representation with only two tokens from scratch. 3) and 4) Multiple \textit{refinement} Tokens: Iteratively predicting \codebracket{$\tokenR$} and adjusting the hand pose step by step demonstrates that performance deteriorates when the number of \codebracket{$\tokenR$} exceeds one. Moreover, we empirically find that predicting more tokens increases the complexity of MLLM training. 5) Single VQ-VAE: We train a single VQ-VAE to predict three grasp tokens simultaneously with a shared codebook. A single network struggles to encapsulate the intricate grasp representation. 6) \textit{w/o} semantic. In this setting, we still use the hierarchical VQ-VAE to predict three tokens, but we do not assign semantic meaning to the tokens. This setup results in decreased performance.\del{6) Separated VQ-VAE: Contrasting with our hierarchical VQ-VAE approach, we tested separate VQ-VAEs for each token. This setup resulted in decreased performance.}

\cheading{Ablation on VQ-VAE settings}
Our investigation into the configurations of VQ-VAE focuses on two aspects: 1) Codebook $\mathcal{B}$ Setting: The size of trainable parameters in the codebook has a significant impact on network performance, leading to either non-convergence or underfitting (see \cref{table:vq_ablation}). 2) Training Strategy: VQ-VAE often suffers from codebook collapse. While methods like exponential moving average (EMA) and codebook reset (Reset) are used to mitigate this (as in \cite{razavi2019generating, zhang2023t2m}), we find that these strategies weaken the representation effectiveness in the grasp representation task. Thus, we opt not to use the EMA strategy and allow each entry to be reset only once during training.

\begin{table}[tb]
    \caption{Ablation study of our MLLM-based grasp generation.}
    \label{tab:mllm_ablations}
    \vspace{-0.5em}
    \centering
    \setlength{\tabcolsep}{5pt}{
        {\scriptsize
        \begin{tabular}{@{}l|c|c|c|c|c@{}}
            \toprule        
                            & P-FID $\downarrow$ & PD $\downarrow$ & SIV $\downarrow$  & GPT-4 $\uparrow$ & PS $\uparrow$ \\
            \midrule
            w/ Llama &   2.38    &   0.51    &   4.20  &  58.9   &   3.8\\
            w/o $\texttt{\codebracket{SO}}$ &   3.74    &   0.70    &    8.20     &   43.3    &   3.2\\
            w/o 2-stage &   4.54   &   0.49    &   5.26    &   62.5   &   4.0\\
            LoRA $r = 16$ &   3.76   &   0.48    &   4.38    &   69.2   &   3.3\\
            LoRA $r = 128$ &   2.68   &   0.50    &   5.10    &   74.5   &   4.0\\
            \bottomrule
        \end{tabular}
        \vspace{-1em}
        }
    }
\end{table}

\cheading{Ablation on MLLM settings}
We conduct ablation studies on the MLLM configurations as presented in \cref{tab:mllm_ablations}:
1) Pretrained LLM: The comparison between Llama-7B \cite{touvron2023llama} and Vicuna-7B models shows Vicuna-7B as more aligned with our grasp generation needs, offering better task suitability. 2) Object size token $\texttt{\codebracket{SO}}$. Contrary to the original ULIP's size normalization of the object point cloud, our findings highlight the significance of object size for grasp generation tasks. 3)\del{ LoRA Rank: We explore the impact of the LoRA rank on the MLLM. We find that a higher rank improves the MLLM's performance, but the improvement is marginal. 4)} Training Stages: Comparing the MLLM trained in a two-stage process with a single-stage approach shows that the former not only enhances effectiveness but also stabilizes the training process. 4) and 5) LoRA rank: We explore the impact of the LoRA rank on the MLLM. We experimentally find that rank 64 is the optimal choice for our task.

\section{Applications}
\label{sec:application}
\setcounter{footnote}{0}
To demonstrate the real-world applicability of the grasps generated by \method, we conducted two case studies in the fields of AR/VR and robotics to show that combined with the RL-based policies, our method can synthesize dynamic grasp motions.

\subsection{Application in AR/VR}
In the context of AR/VR, producing grasps that align with user intent and facilitate natural object manipulation is essential. We evaluated the practicality of grasps generated by \method using the D-grasp method \cite{christen2022d} within the RaiSim \cite{hwangbo2018per} simulated environment. D-grasp, which is based on a reinforcement learning (RL) approach, focuses on creating dynamic human grasps. It calculates the next grasp action from a static reference grasp $\bar{\grasp}$, the target object position $\bar{\mbold{T}}_{\obj}$, and the current state, including the hand and object's pose and velocity, using a policy $\mbold{\pi}$ trained with the\del{ Proximal Policy Optimization} PPO \cite{schulman2017proximal} algorithm.

For our experiments, \method generates the reference pose $\bar{\grasp}$ for a specified object $\obj$ and language instruction $\lang$. We utilize the publicly available D-grasp checkpoints\footnote{\url{https://github.com/christsa/dgrasp}} to synthesize dynamic grasps. The object is targeted to lift 10 $cm$ upwards along the gravitational direction. We maintain the hand shape parameter $\mbold{\beta}$ at $\mbold{0}$, consistent with D-grasp. This process is illustrated in \cref{fig:application-arvr}.\del{ For further details and results, please refer to \supp.}

\subsection{Application in Embodied Robotics}

Our method's efficacy is further validated in embodied robotics. Following the UniDexGrasp\footnote{\url{https://github.com/PKU-EPIC/UniDexGrasp}} methodology \cite{xu2023unidexgrasp}, which entails static reference grasp generation and goal-conditioned grasp execution, our experiments focus on the latter phase to assess the performance of \method-generated grasps. We retarget our generated grasp $\bar{\grasp}$ to the ShadowHand $\mathcal{S}$ \cite{shadowrobot} and evaluate the grasp within the \del{GPU-based }IsaacGym simulation\del{ environment}\cite{makoviychuk2021isaac}.

\begin{figure}[tb]
    \centering
    \begin{subfigure}{0.48\linewidth}
      \includegraphics[width=\linewidth]{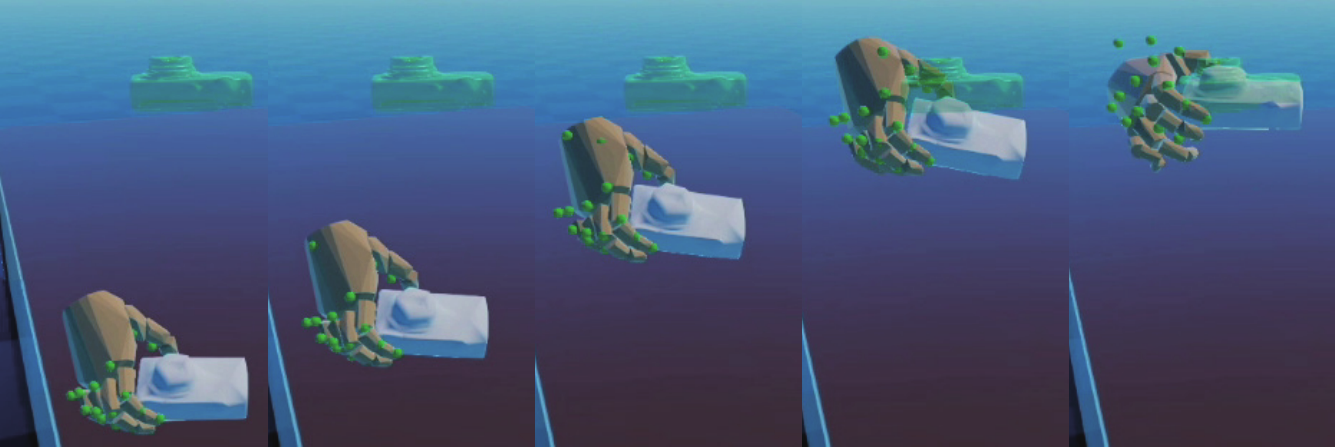}
      \caption{Human-like grasps motion in AR/VR}
      \label{fig:application-arvr}
    \end{subfigure}
    \hfill
    \begin{subfigure}{0.48\linewidth}
      \includegraphics[width=\linewidth]{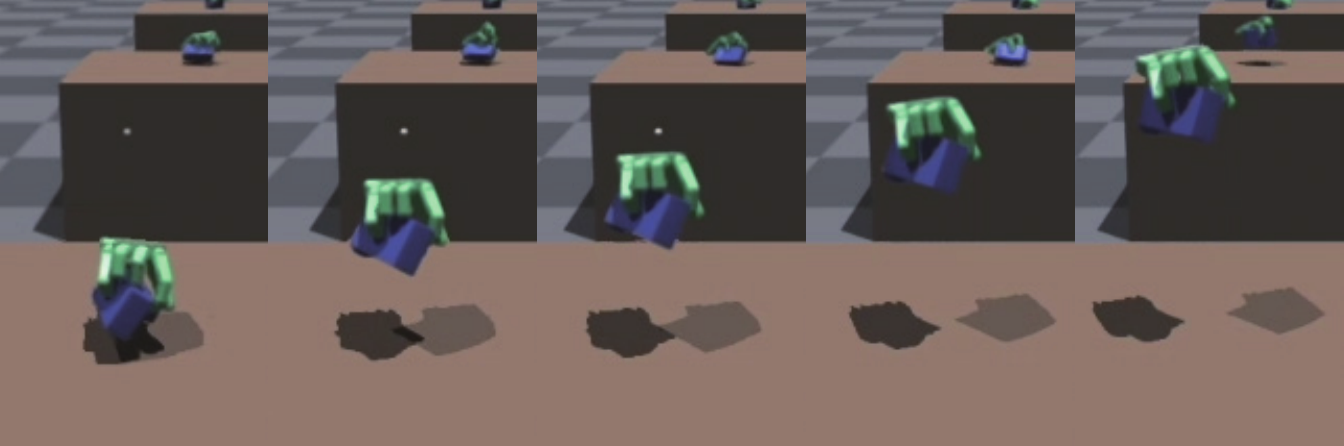}
      \caption{Dexterous grasps execution in robotics}
      \label{fig:application-robotics}
    \end{subfigure}
    \caption{Applications of our \method~in both the AR/VR environment and robotics scenario.}
    \label{fig:short}
\end{figure}

Initially, \method produces the reference grasp pose $\bar{\grasp}$. We then establish a pipeline to adapt $\bar{\grasp}$ for the ShadowHand model $\bar{\grasp}_\mathcal{S}$, beginning with aligning finger keypoints between the MANO and ShadowHand models. Given the significant differences in degrees of freedom (DoF) and morphology between the two models, we implement a fitting-based optimization approach to refine the ShadowHand-object interaction. After optimizing $\bar{\grasp}_\mathcal{S}$, we apply the off-the-shelf UniDexGrasp's pretrained policy to execute the dynamic grasp sequence. Details on our fitting pipeline and the integration of $\bar{\grasp}_\mathcal{S}$ with UniDexGrasp are further elaborated in the \supp. \cref{fig:application-robotics} showcases the process of grasping and lifting as generated by UniDexGrasp.

\section{Conclusion}\label{sec:conclusion}
\del{We introduce \method, an approach aimed at generating semantic grasps from language instructions. We propose a novel grasp representation that emulates the natural human grasping process. The discretized representation is both interpretable and controllable, making it ideal for semantic space alignment. Leveraging this representation, we deploy MLLM to generate grasps from language instructions. Tailored for this task, we also present \dataset, a comprehensive dataset containing grasp-text-aligned annotations. As we explore potential applications in AR/VR and embodied robotics, we are hopeful that \method will contribute to advancements in generating more human-like grasps in various contexts.}

We introduce \method, an approach aimed at generating semantic grasps from language instructions. We propose a novel grasp representation that emulates the natural human grasping process. The discretized representation is both interpretable and controllable, making it ideal for semantic space alignment. Leveraging this representation, we deploy MLLM to generate grasps from language instructions. Tailored for this task, we also present \dataset, a comprehensive dataset containing grasp-text-aligned annotations. As we explore potential applications in AR/VR and embodied robotics, we are hopeful that \method will contribute to advancements in generating more human-like, semantically coherent grasps in various contexts.

\cheading{Limitations} Despite \method demonstrating the capability to generate static single-hand grasps from semantic cues and dynamic grasps through RL integration, exploration remains in two directions: two-hand manipulation and end-to-end semantic grasp motion synthesis. The former requires addressing both hands' cooperation, and the latter, the continuity of motion, both contingent on the availability of extensive, high-quality motion capture or synthesis data for training.\del{ These two directions are valuable, and we aim to explore them in future work.} Tackling these challenges promises to advance embodied grasping, pushing toward more sophisticated and realistic manipulation.


%
%
\bibliographystyle{splncs04}
\bibliography{main}

\begin{thebibliography}{10}
\providecommand{\url}[1]{\texttt{#1}}
\providecommand{\urlprefix}{URL }
\providecommand{\doi}[1]{https://doi.org/#1}

\bibitem{bohg2013data}
Bohg, J., Morales, A., Asfour, T., Kragic, D.: Data-driven grasp synthesis—a survey. IEEE Transactions on robotics  (2013)

\bibitem{brahmbhatt2019contactgrasp}
Brahmbhatt, S., Handa, A., Hays, J., Fox, D.: Contactgrasp: Functional multi-finger grasp synthesis from contact. In: IROS (2019)

\bibitem{brahmbhatt2020contactpose}
Brahmbhatt, S., Tang, C., Twigg, C.D., Kemp, C.C., Hays, J.: Contactpose: A dataset of grasps with object contact and hand pose. In: ECCV (2020)

\bibitem{chao2021dexycb}
Chao, Y.W., Yang, W., Xiang, Y., Molchanov, P., Handa, A., Tremblay, J., Narang, Y.S., Van~Wyk, K., Iqbal, U., Birchfield, S., et~al.: Dexycb: A benchmark for capturing hand grasping of objects. In: CVPR (2021)

\bibitem{christen2022d}
Christen, S., Kocabas, M., Aksan, E., Hwangbo, J., Song, J., Hilliges, O.: D-grasp: Physically plausible dynamic grasp synthesis for hand-object interactions. In: CVPR (2022)

\bibitem{corona2020ganhand}
Corona, E., Pumarola, A., Alenya, G., Moreno-Noguer, F., Rogez, G.: Ganhand: Predicting human grasp affordances in multi-object scenes. In: CVPR (2020)

\bibitem{coumanspybullet}
Coumans, E., Bai, Y.: Pybullet, a python module for physics simulation in robotics, games and machine learning. URL http://pybullet. org  (2017)

\bibitem{devlin2018bert}
Devlin, J., Chang, M.W., Lee, K., Toutanova, K.: Bert: Pre-training of deep bidirectional transformers for language understanding. arXiv preprint arXiv:1810.04805  (2018)

\bibitem{eppner2021acronym}
Eppner, C., Mousavian, A., Fox, D.: Acronym: A large-scale grasp dataset based on simulation. In: ICRA (2021)

\bibitem{fan2023arctic}
Fan, Z., Taheri, O., Tzionas, D., Kocabas, M., Kaufmann, M., Black, M.J., Hilliges, O.: Arctic: A dataset for dexterous bimanual hand-object manipulation. In: CVPR (2023)

\bibitem{fang2023anygrasp}
Fang, H.S., Wang, C., Fang, H., Gou, M., Liu, J., Yan, H., Liu, W., Xie, Y., Lu, C.: Anygrasp: Robust and efficient grasp perception in spatial and temporal domains. IEEE Transactions on Robotics  (2023)

\bibitem{fang2020graspnet}
Fang, H.S., Wang, C., Gou, M., Lu, C.: Graspnet-1billion: A large-scale benchmark for general object grasping. In: CVPR (2020)

\bibitem{feix2015grasp}
Feix, T., Romero, J., Schmiedmayer, H.B., Dollar, A.M., Kragic, D.: The grasp taxonomy of human grasp types. IEEE Transactions on human-machine systems  (2015)

\bibitem{feng2023posegpt}
Feng, Y., Lin, J., Dwivedi, S.K., Sun, Y., Patel, P., Black, M.J.: Posegpt: Chatting about 3d human pose. arXiv preprint arXiv:2311.18836  (2023)

\bibitem{gao2022dart}
Gao, D., Xiu, Y., Li, K., Yang, L., Wang, F., Zhang, P., Zhang, B., Lu, C., Tan, P.: Dart: Articulated hand model with diverse accessories and rich textures. NeurIPS  (2022)

\bibitem{garcia2018first}
Garcia-Hernando, G., Yuan, S., Baek, S., Kim, T.K.: First-person hand action benchmark with rgb-d videos and 3d hand pose annotations. In: CVPR (2018)

\bibitem{gemini}
{Gemini}: {Introduction to Gemini}. \url{https://deepmind.google/technologies/gemini/\#introduction} (2023)

\bibitem{goodfellow2014generative}
Goodfellow, I., Pouget-Abadie, J., Mirza, M., Xu, B., Warde-Farley, D., Ozair, S., Courville, A., Bengio, Y.: Generative adversarial nets. NeurIPS  (2014)

\bibitem{grady2021contactopt}
Grady, P., Tang, C., Twigg, C.D., Vo, M., Brahmbhatt, S., Kemp, C.C.: Contactopt: Optimizing contact to improve grasps. In: CVPR (2021)

\bibitem{hampali2020honnotate}
Hampali, S., Rad, M., Oberweger, M., Lepetit, V.: Honnotate: A method for 3d annotation of hand and object poses. In: CVPR (2020)

\bibitem{hampali2022keypoint}
Hampali, S., Sarkar, S.D., Rad, M., Lepetit, V.: Keypoint transformer: Solving joint identification in challenging hands and object interactions for accurate 3d pose estimation. In: CVPR (2022)

\bibitem{hasson2019learning}
Hasson, Y., Varol, G., Tzionas, D., Kalevatykh, I., Black, M.J., Laptev, I., Schmid, C.: Learning joint reconstruction of hands and manipulated objects. In: CVPR (2019)

\bibitem{hu2021lora}
Hu, E.J., Shen, Y., Wallis, P., Allen-Zhu, Z., Li, Y., Wang, S., Wang, L., Chen, W.: Lora: Low-rank adaptation of large language models. arXiv preprint arXiv:2106.09685  (2021)

\bibitem{huang2023embodied}
Huang, J., Yong, S., Ma, X., Linghu, X., Li, P., Wang, Y., Li, Q., Zhu, S.C., Jia, B., Huang, S.: An embodied generalist agent in 3d world. arXiv preprint arXiv:2311.12871  (2023)

\bibitem{huang2020manifoldplus}
Huang, J., Zhou, Y., Guibas, L.: Manifoldplus: A robust and scalable watertight manifold surface generation method for triangle soups. arXiv preprint arXiv:2005.11621  (2020)

\bibitem{huang2023language}
Huang, S., Dong, L., Wang, W., Hao, Y., Singhal, S., Ma, S., Lv, T., Cui, L., Mohammed, O.K., Liu, Q., et~al.: Language is not all you need: Aligning perception with language models. arXiv preprint arXiv:2302.14045  (2023)

\bibitem{hwangbo2018per}
Hwangbo, J., Lee, J., Hutter, M.: Per-contact iteration method for solving contact dynamics. IEEE Robotics and Automation Letters  (2018)

\bibitem{jian2023affordpose}
Jian, J., Liu, X., Li, M., Hu, R., Liu, J.: Affordpose: A large-scale dataset of hand-object interactions with affordance-driven hand pose. In: ICCV (2023)

\bibitem{jiang2023motiongpt}
Jiang, B., Chen, X., Liu, W., Yu, J., Yu, G., Chen, T.: Motiongpt: Human motion as a foreign language. NeurIPS  (2023)

\bibitem{jiang2021hand}
Jiang, H., Liu, S., Wang, J., Wang, X.: Hand-object contact consistency reasoning for human grasps generation. In: ICCV (2021)

\bibitem{jin2024reasoning}
Jin, S., Xu, J., Lei, Y., Zhang, L.: Reasoning grasping via multimodal large language model. arXiv preprint arXiv:2402.06798  (2024)

\bibitem{karunratanakul2021skeleton}
Karunratanakul, K., Spurr, A., Fan, Z., Hilliges, O., Tang, S.: A skeleton-driven neural occupancy representation for articulated hands. In: 3DV (2021)

\bibitem{karunratanakul2020grasping}
Karunratanakul, K., Yang, J., Zhang, Y., Black, M.J., Muandet, K., Tang, S.: Grasping field: Learning implicit representations for human grasps. In: 3DV (2020)

\bibitem{kim2024parahome}
Kim, J., Kim, J., Na, J., Joo, H.: Parahome: Parameterizing everyday home activities towards 3d generative modeling of human-object interactions. arXiv preprint arXiv:2401.10232  (2024)

\bibitem{kudo2018sentencepiece}
Kudo, T., Richardson, J.: Sentencepiece: A simple and language independent subword tokenizer and detokenizer for neural text processing. arXiv preprint arXiv:1808.06226  (2018)

\bibitem{kwon2021h2o}
Kwon, T., Tekin, B., St{\"u}hmer, J., Bogo, F., Pollefeys, M.: H2o: Two hands manipulating objects for first person interaction recognition. In: ICCV (2021)

\bibitem{lakshmipathy2023contact}
Lakshmipathy, A.S., Feng, N., Lee, Y.X., Mahler, M., Pollard, N.: Contact edit: Artist tools for intuitive modeling of hand-object interactions. ACM TOG  (2023)

\bibitem{li2022contact2grasp}
Li, H., Lin, X., Zhou, Y., Li, X., Huo, Y., Chen, J., Ye, Q.: Contact2grasp: 3d grasp synthesis via hand-object contact constraint. In: IJCAI (2023)

\bibitem{Li_Yang_Lin_Xu_Zhan_Zhao_Zhu_Kang_Wu_Lu_2024}
Li, K., Yang, L., Lin, Z., Xu, J., Zhan, X., Zhao, Y., Zhu, P., Kang, W., Wu, K., Lu, C.: Favor: Full-body ar-driven virtual object rearrangement guided by instruction text. Proceedings of the AAAI Conference on Artificial Intelligence  (2024)

\bibitem{li2023chord}
Li, K., Yang, L., Zhen, H., Lin, Z., Zhan, X., Zhong, L., Xu, J., Wu, K., Lu, C.: Chord: Category-level hand-held object reconstruction via shape deformation. In: ICCV (2023)

\bibitem{li2023videochat}
Li, K., He, Y., Wang, Y., Li, Y., Wang, W., Luo, P., Wang, Y., Wang, L., Qiao, Y.: Videochat: Chat-centric video understanding. arXiv preprint arXiv:2305.06355  (2023)

\bibitem{li2022nimble}
Li, Y., Zhang, L., Qiu, Z., Jiang, Y., Li, N., Ma, Y., Zhang, Y., Xu, L., Yu, J.: Nimble: a non-rigid hand model with bones and muscles. ACM TOG  (2022)

\bibitem{liu2023llava}
Liu, H., Li, C., Wu, Q., Lee, Y.J.: Visual instruction tuning. NeurIPS  (2023)

\bibitem{liu2020deep}
Liu, M., Pan, Z., Xu, K., Ganguly, K., Manocha, D.: Deep differentiable grasp planner for high-dof grippers. arXiv preprint arXiv:2002.01530  (2020)

\bibitem{liu2023contactgen}
Liu, S., Zhou, Y., Yang, J., Gupta, S., Wang, S.: Contactgen: Generative contact modeling for grasp generation. In: ICCV (2023)

\bibitem{liu2021synthesizing}
Liu, T., Liu, Z., Jiao, Z., Zhu, Y., Zhu, S.C.: Synthesizing diverse and physically stable grasps with arbitrary hand structures using differentiable force closure estimator. IEEE Robotics and Automation Letters  (2021)

\bibitem{liu2024realdex}
Liu, Y., Yang, Y., Wang, Y., Wu, X., Wang, J., Yao, Y., Schwertfeger, S., Yang, S., Wang, W., Yu, J., et~al.: Realdex: Towards human-like grasping for robotic dexterous hand. arXiv preprint arXiv:2402.13853  (2024)

\bibitem{liu2024taco}
Liu, Y., Yang, H., Si, X., Liu, L., Li, Z., Zhang, Y., Liu, Y., Yi, L.: Taco: Benchmarking generalizable bimanual tool-action-object understanding. arXiv preprint arXiv:2401.08399  (2024)

\bibitem{liu2022hoi4d}
Liu, Y., Liu, Y., Jiang, C., Lyu, K., Wan, W., Shen, H., Liang, B., Fu, Z., Wang, H., Yi, L.: Hoi4d: A 4d egocentric dataset for category-level human-object interaction. In: CVPR (2022)

\bibitem{lu2023ugg}
Lu, J., Kang, H., Li, H., Liu, B., Yang, Y., Huang, Q., Hua, G.: Ugg: Unified generative grasping. arXiv preprint arXiv:2311.16917  (2023)

\bibitem{makoviychuk2021isaac}
Makoviychuk, V., Wawrzyniak, L., Guo, Y., Lu, M., Storey, K., Macklin, M., Hoeller, D., Rudin, N., Allshire, A., Handa, A., et~al.: Isaac gym: High performance gpu-based physics simulation for robot learning. arXiv preprint arXiv:2108.10470  (2021)

\bibitem{miller2004graspit}
Miller, A.T., Allen, P.K.: Graspit! a versatile simulator for robotic grasping. IEEE Robotics \& Automation Magazine  (2004)

\bibitem{mousavian20196}
Mousavian, A., Eppner, C., Fox, D.: 6-dof graspnet: Variational grasp generation for object manipulation. In: ICCV (2019)

\bibitem{newbury2023deep}
Newbury, R., Gu, M., Chumbley, L., Mousavian, A., Eppner, C., Leitner, J., Bohg, J., Morales, A., Asfour, T., Kragic, D., et~al.: Deep learning approaches to grasp synthesis: A review. IEEE Transactions on Robotics  (2023)

\bibitem{nichol2022point}
Nichol, A., Jun, H., Dhariwal, P., Mishkin, P., Chen, M.: Point-e: A system for generating 3d point clouds from complex prompts. arXiv preprint arXiv:2212.08751  (2022)

\bibitem{openai2023gpt}
OpenAI: Gpt-4 technical report. arXiv preprint arXiv:2303.08774  (2023)

\bibitem{GPT4Vision23}
OpenAI: Gpt-4v(ision) system card (2023), \url{https://openai.com/research/gpt-4v-system-card}

\bibitem{qin2022dexmv}
Qin, Y., Wu, Y.H., Liu, S., Jiang, H., Yang, R., Fu, Y., Wang, X.: Dexmv: Imitation learning for dexterous manipulation from human videos. In: ECCV (2022)

\bibitem{radford2021learning}
Radford, A., Kim, J.W., Hallacy, C., Ramesh, A., Goh, G., Agarwal, S., Sastry, G., Askell, A., Mishkin, P., Clark, J., et~al.: Learning transferable visual models from natural language supervision. In: ICML (2021)

\bibitem{raffel2020exploring}
Raffel, C., Shazeer, N., Roberts, A., Lee, K., Narang, S., Matena, M., Zhou, Y., Li, W., Liu, P.J.: Exploring the limits of transfer learning with a unified text-to-text transformer. The Journal of Machine Learning Research  (2020)

\bibitem{razavi2019generating}
Razavi, A., Van~den Oord, A., Vinyals, O.: Generating diverse high-fidelity images with vq-vae-2. NeurIPS  (2019)

\bibitem{MANO:SIGGRAPHASIA:2017}
Romero, J., Tzionas, D., Black, M.J.: Embodied hands: Modeling and capturing hands and bodies together. ACM TOG  (2017)

\bibitem{schulman2017proximal}
Schulman, J., Wolski, F., Dhariwal, P., Radford, A., Klimov, O.: Proximal policy optimization algorithms. arXiv preprint arXiv:1707.06347  (2017)

\bibitem{sener2022assembly101}
Sener, F., Chatterjee, D., Shelepov, D., He, K., Singhania, D., Wang, R., Yao, A.: Assembly101: A large-scale multi-view video dataset for understanding procedural activities. In: CVPR (2022)

\bibitem{shadowrobot}
{Shadowrobot}: {Dexterous Hand Series}. \url{https://www.shadowrobot.com/dexterous-hand-series/} (2005)

\bibitem{sohn2015learning}
Sohn, K., Lee, H., Yan, X.: Learning structured output representation using deep conditional generative models. NeurIPS  (2015)

\bibitem{taheri2022goal}
Taheri, O., Choutas, V., Black, M.J., Tzionas, D.: Goal: Generating 4d whole-body motion for hand-object grasping. In: CVPR (2022)

\bibitem{taheri2020grab}
Taheri, O., Ghorbani, N., Black, M.J., Tzionas, D.: Grab: A dataset of whole-body human grasping of objects. In: ECCV (2020)

\bibitem{tang2023graspgpt}
Tang, C., Huang, D., Ge, W., Liu, W., Zhang, H.: Graspgpt: Leveraging semantic knowledge from a large language model for task-oriented grasping. IEEE Robotics and Automation Letters  (2023)

\bibitem{touvron2023llama}
Touvron, H., Martin, L., Stone, K., Albert, P., Almahairi, A., Babaei, Y., Bashlykov, N., Batra, S., Bhargava, P., Bhosale, S., et~al.: Llama 2: Open foundation and fine-tuned chat models. arXiv preprint arXiv:2307.09288  (2023)

\bibitem{turpin2022grasp}
Turpin, D., Wang, L., Heiden, E., Chen, Y.C., Macklin, M., Tsogkas, S., Dickinson, S., Garg, A.: Grasp’d: Differentiable contact-rich grasp synthesis for multi-fingered hands. In: ECCV (2022)

\bibitem{van2017neural}
Van Den~Oord, A., Vinyals, O., et~al.: Neural discrete representation learning. NeurIPS  (2017)

\bibitem{wan2023unidexgrasp++}
Wan, W., Geng, H., Liu, Y., Shan, Z., Yang, Y., Yi, L., Wang, H.: Unidexgrasp++: Improving dexterous grasping policy learning via geometry-aware curriculum and iterative generalist-specialist learning. In: ICCV (2023)

\bibitem{wei2022approximate}
Wei, X., Liu, M., Ling, Z., Su, H.: Approximate convex decomposition for 3d meshes with collision-aware concavity and tree search. ACM TOG  (2022)

\bibitem{wu2022saga}
Wu, Y., Wang, J., Zhang, Y., Zhang, S., Hilliges, O., Yu, F., Tang, S.: Saga: Stochastic whole-body grasping with contact. In: ECCV (2022)

\bibitem{xie2023hmdo}
Xie, W., Yu, Z., Zhao, Z., Zuo, B., Wang, Y.: Hmdo: Markerless multi-view hand manipulation capture with deformable objects. Graphical Models  (2023)

\bibitem{xu2023pointllm}
Xu, R., Wang, X., Wang, T., Chen, Y., Pang, J., Lin, D.: Pointllm: Empowering large language models to understand point clouds. arXiv preprint arXiv:2308.16911  (2023)

\bibitem{xu2023unidexgrasp}
Xu, Y., Wan, W., Zhang, J., Liu, H., Shan, Z., Shen, H., Wang, R., Geng, H., Weng, Y., Chen, J., et~al.: Unidexgrasp: Universal robotic dexterous grasping via learning diverse proposal generation and goal-conditioned policy. In: CVPR (2023)

\bibitem{xue2023ulip}
Xue, L., Yu, N., Zhang, S., Li, J., Mart{\'\i}n-Mart{\'\i}n, R., Wu, J., Xiong, C., Xu, R., Niebles, J.C., Savarese, S.: Ulip-2: Towards scalable multimodal pre-training for 3d understanding. arXiv preprint arXiv:2305.08275  (2023)

\bibitem{yang2022artiboost}
Yang, L., Li, K., Zhan, X., Lv, J., Xu, W., Li, J., Lu, C.: Artiboost: Boosting articulated 3d hand-object pose estimation via online exploration and synthesis. In: CVPR (2022)

\bibitem{yang2022oakink}
Yang, L., Li, K., Zhan, X., Wu, F., Xu, A., Liu, L., Lu, C.: Oakink: A large-scale knowledge repository for understanding hand-object interaction. In: CVPR (2022)

\bibitem{yang2024learning}
Yang, L., Zhan, X., Li, K., Xu, W., Zhang, J., Li, J., Lu, C.: Learning a contact potential field for modeling the hand-object interaction. IEEE Transactions on Pattern Analysis and Machine Intelligence  (2024)

\bibitem{yin2023shapegpt}
Yin, F., Chen, X., Zhang, C., Jiang, B., Zhao, Z., Fan, J., Yu, G., Li, T., Chen, T.: Shapegpt: 3d shape generation with a unified multi-modal language model. arXiv preprint arXiv:2311.17618  (2023)

\bibitem{yu2022point}
Yu, X., Tang, L., Rao, Y., Huang, T., Zhou, J., Lu, J.: Point-bert: Pre-training 3d point cloud transformers with masked point modeling. In: CVPR (2022)

\bibitem{zhan2024oakink2}
Zhan, X., Yang, L., Zhao, Y., Mao, K., Xu, H., Lin, Z., Li, K., Lu, C.: Oakink2: A dataset of bimanual hands-object manipulation in complex task completion. arXiv preprint arXiv:2403.19417  (2024)

\bibitem{zhang2023video}
Zhang, H., Li, X., Bing, L.: Video-llama: An instruction-tuned audio-visual language model for video understanding. arXiv preprint arXiv:2306.02858  (2023)

\bibitem{zhang2021manipnet}
Zhang, H., Ye, Y., Shiratori, T., Komura, T.: Manipnet: neural manipulation synthesis with a hand-object spatial representation. ACM TOG  (2021)

\bibitem{zhang2023t2m}
Zhang, J., Zhang, Y., Cun, X., Huang, S., Zhang, Y., Zhao, H., Lu, H., Shen, X.: T2m-gpt: Generating human motion from textual descriptions with discrete representations. In: CVPR (2023)

\bibitem{zheng2023cams}
Zheng, J., Zheng, Q., Fang, L., Liu, Y., Yi, L.: Cams: Canonicalized manipulation spaces for category-level functional hand-object manipulation synthesis. In: CVPR (2023)

\bibitem{zheng2023judging}
Zheng, L., Chiang, W.L., Sheng, Y., Zhuang, S., Wu, Z., Zhuang, Y., Lin, Z., Li, Z., Li, D., Xing, E.P., Zhang, H., Gonzalez, J.E., Stoica, I.: Judging llm-as-a-judge with mt-bench and chatbot arena. arXiv preprint arXiv:2306.05685  (2023)

\bibitem{zhou2019continuity}
Zhou, Y., Barnes, C., Lu, J., Yang, J., Li, H.: On the continuity of rotation representations in neural networks. In: CVPR (2019)

\bibitem{zhu2023minigpt}
Zhu, D., Chen, J., Shen, X., Li, X., Elhoseiny, M.: Minigpt-4: Enhancing vision-language understanding with advanced large language models. arXiv preprint arXiv:2304.10592  (2023)

\bibitem{zhu2021toward}
Zhu, T., Wu, R., Lin, X., Sun, Y.: Toward human-like grasp: Dexterous grasping via semantic representation of object-hand. In: ICCV (2021)

\bibitem{zhu2023contactart}
Zhu, Z., Wang, J., Qin, Y., Sun, D., Jampani, V., Wang, X.: Contactart: Learning 3d interaction priors for category-level articulated object and hand poses estimation. arXiv preprint arXiv:2305.01618  (2023)

\end{thebibliography}

\clearpage

\begin{appendix}
    \chapter*{Appendices}
    \addcontentsline{toc}{chapter}{Appendices}
    \section{Experiments Details}
\label{sec:supp_ablation}

\subsection{Setting Details}

\cheading{Dataset Split} Our dataset, \dataset, builds upon and extends the OakInk dataset \cite{yang2022oakink}. As such, we adopt the same split as OakInk, with 80\% of the data allocated for training, 10\% for validation, and 10\% for testing. We ensure that the test set includes a wide variety of objects and language instructions, thereby allowing us to evaluate the generalization capabilities of our \method.

\cheading{MLLM prompt} The prompt for our grasp-aware MLLM is specified in \cref{table:supp_prompt_mllm}, guiding the model to generate coherent and plausible grasps from language instructions.

\cheading{GPT-4 assisted evaluation} We leverage the commercial GPT-4v \cite{GPT4Vision23} to evaluate the quality of our \method. This involves rendered images $\mbold{I}$ of the generated grasps $\hat{\grasp}$, alongside the evaluation prompt outlined in \cref{table:supp_prompt_eval} to GPT-4v. The model then returns a quality score reflecting both semantic similarity and physical reliability. To enhance the accuracy of GPT-4v, we take the following methods to improve the quality of the rendered images: 1) The grasp $\hat{\grasp}$ is mapped onto the differentiable NIMBLE model \cite{li2022nimble}, which contains delicate muscle modeling and high-fidelity hand skin textures. 2) Images are rendered in Blender using the Cycles rendering engine, complemented by random lighting and camera positioning to ensure diversity.

\cheading{Perceptual Score} We ask 5 volunteers to rate the quality of the generated grasps $\hat{\grasp}$ on a 5-point Likert scale. We randomly sample 50 predicated grasps from the test set for each experiment. The evaluation indicators involve the following three aspects: 1) Semantic coherence with the provided language instructions, 2) Physical plausibility of the hand pose, and 3) Stability of the interaction between the hand and the object. The perceptual score is the average of the ratings.

\subsection{Representation Ablation Studies Details}

This section elaborates on the ablation studies conducted to examine our discrete representation.

\cheading{Single Token} Contrary to our primary model's multi-token and hierarchical VQ-VAE structure, we explore a simplified model using a single VQ-VAE with one codebook to encapsulate the entire grasp representation.\del{The codebook's dimension, $d_\mathcal{B}$, is maintained at 256, with $K = 512$ entries, mirroring the specifications of our main model.}

\cheading{The \codebracket{$\tokenO, \tokenM$} Setting} In this variant, we devise a dual-layer hierarchical VQ-VAE specifically for grasp representation that is trained from scratch. The first codebook is for the \textit{orientation} and the second codebook is for the \textit{manner}.

\cheading{Multiple \textit{refinement} Tokens} This configuration introduces a delta VQ-VAE designed to refine the grasp pose by predicting incremental \textit{refinement} tokens \codebracket{$\tokenR$} conditioned on the preceding hand grasp and object point cloud. Based on this setting, we can iteratively adjust the hand pose by applying the delta parameters to the previous grasp.

\cheading{Single VQ-VAE} Here, a unified VQ-VAE codebook is employed to simultaneously derive the three tokens (\codebracket{$\tokenO, \tokenM, \tokenR$}), each decoded into the target pose through distinct decoding head.

\cheading{Without Semantic} In our primary model, the reconstruction loss is composed of three parts: the \textit{orientation} loss $\mathcal{L}_{\tokenO}$, the \textit{manner} loss $\mathcal{L}_{\tokenM}$, and the \textit{refinement} loss $\mathcal{L}_{\tokenR}$. Specifically, $\mathcal{L}_{\tokenO}$ only supervises the $\hat{\mbold{T}}$ and $\mathcal{L}_{\tokenM}$ mainly focuses on the $\hat{\mbold{\theta}}, \hat{\mbold{\beta}}$. We simply set the items that are not supervised to zero in the loss function \cref{eq:supp_loss}.

\begin{align}
    \mathcal{L}_{\text{rec}} =&\quad\mathcal{L}_{\tokenO} + \mathcal{L}_{\tokenO} + \mathcal{L}_{\tokenR} \notag\\
    = &\quad \|\mathcal{M}(\mbold{T}, \mbold{0}, \mbold{0}) - \mathcal{M}(\hat{\mbold{T}}, \mbold{0}, \mbold{0})\|_2^2 \notag \\
    &+ \|\mathcal{M}(\mbold{T}, \mbold{\theta}, \mbold{\beta}) - \mathcal{M}(\hat{\mbold{T}}, \hat{\mbold{\theta}}, \hat{\mbold{\beta}})\|_2^2 \notag\\
    &+ \|\mathcal{M}(\mbold{T}, \mbold{\theta}, \mbold{\beta}) - \mathcal{M}(\Delta\hat{\mbold{T}} \cdot \hat{\mbold{T}}, \Delta\hat{\mbold{\theta}} + \hat{\mbold{\theta}}, \Delta\hat{\mbold{\beta}} + \hat{\mbold{\beta}})\|_2^2
    \label{eq:supp_loss}
\end{align}

In the \textit{without semantic} scenario, the \codebracket{$\tokenO$} token is not exclusively constrained to represent \textit{orientation}. We experimentally find that, during training, the three tokens collapse into a single token, which degrades performance.

\begin{figure}[b!]
    \centering
    \includegraphics[width=\linewidth]{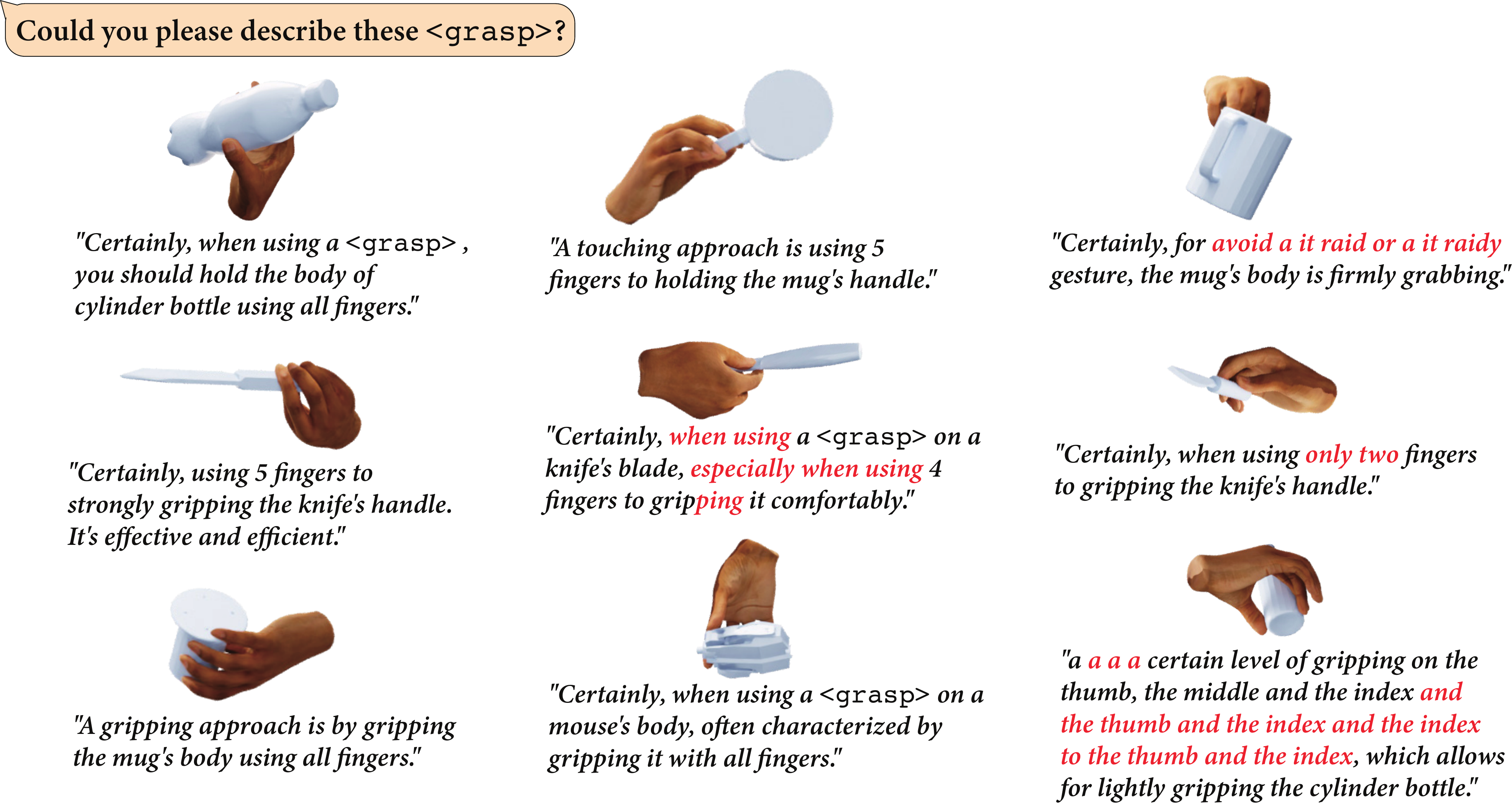}
    \caption{\textbf{Results on the grasp caption task.} The {\color[RGB]{255,0,0} red text} indicates the mistakes or the hallucinations of our model.}
    \label{fig:supp_caption}
\end{figure}

\section{Exploratory Study}
\label{sec:supp_explore}
In our \method, we focus on training a grasp-aware MLLM to synchronize three distinct modalities—grasps, object models, and language instructions—within a unified representational space. We conduct an exploratory study to investigate the effectiveness of the alignment. Our findings indicate that the MLLM is not only capable of generating semantic grasps but also demonstrates promise in the grasp captioning task. Specifically, when provided with grasp tokens, the MLLM is able to produce corresponding language descriptions that capture both the low-level details and high-level intents of the grasps in some instances (as illustrated in \cref{fig:supp_caption}). It is important to admit that these generated language descriptions do not always achieve the same level of accuracy as the ground truth. The MLLM occasionally struggles to capture the details of interactions or to hallucinate details in its language descriptions. This discrepancy underscores the inherent complexity of the caption task, which necessitates a comprehensive understanding of point clouds, intent interpretation, interaction reasoning, and natural language generation capabilities. However, it is a promising direction to explore the potential of the MLLM in the future when scaling up the model and the dataset.

\section{Applications Details}
\label{sec:supp_application}

\begin{figure}[b!]
    \centering
    \includegraphics[width=\linewidth]{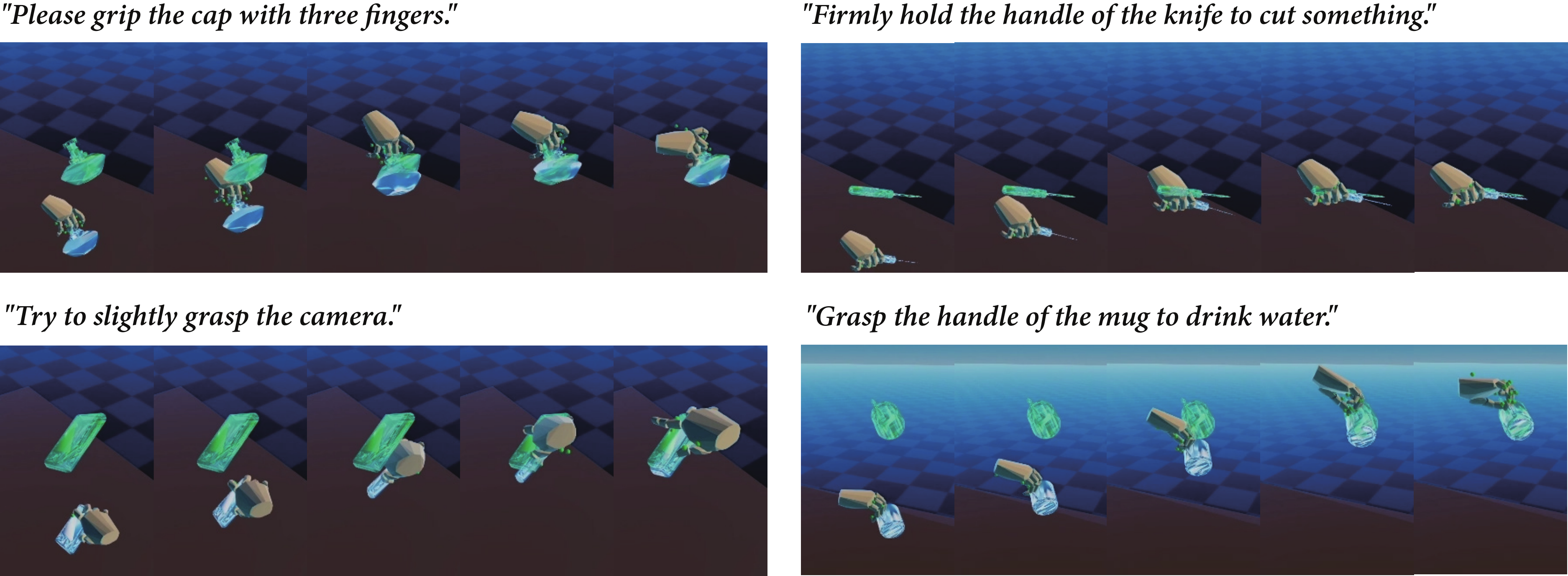}
    \caption{Synthesis of human-like grasps motion in AR/VR application.}
    \label{fig:supp_dgrasp}
\end{figure}

\subsection{Physical-Plausible Dynamic Grasp using Human-like Hand}
In our VR/AR application, we employ the open-source D-grasp method \cite{christen2022d} to synthesize dynamic, human-like grasps. As described in the main paper, the reference pose $\bar{\grasp}$, corresponding to the language instruction $\lang$, is generated using our \method. This dynamic grasp policy is then applied to assess the feasibility of the generated grasps.
To ensure alignment with real-world scenarios, we rotate the hand-object grasp pair so the palm faces toward the table. Given that the OakInk dataset does not provide object weight information, we assign a hypothetical weight of 300g to each object for the purpose of this evaluation. Samples of the generated dynamic grasps are illustrated in \cref{fig:supp_dgrasp}. In our analysis, any relative sliding between the hand and the object exceeding 4cm is classified as a failure. Based on this criterion, we report a success rate of 62.9\% for our generated grasps on the test set.

\subsection{Physical-Plausible Dynamic Grasp using ShaodowHand}
\begin{figure}[t!]
    \centering
    \includegraphics[width=0.93\linewidth]{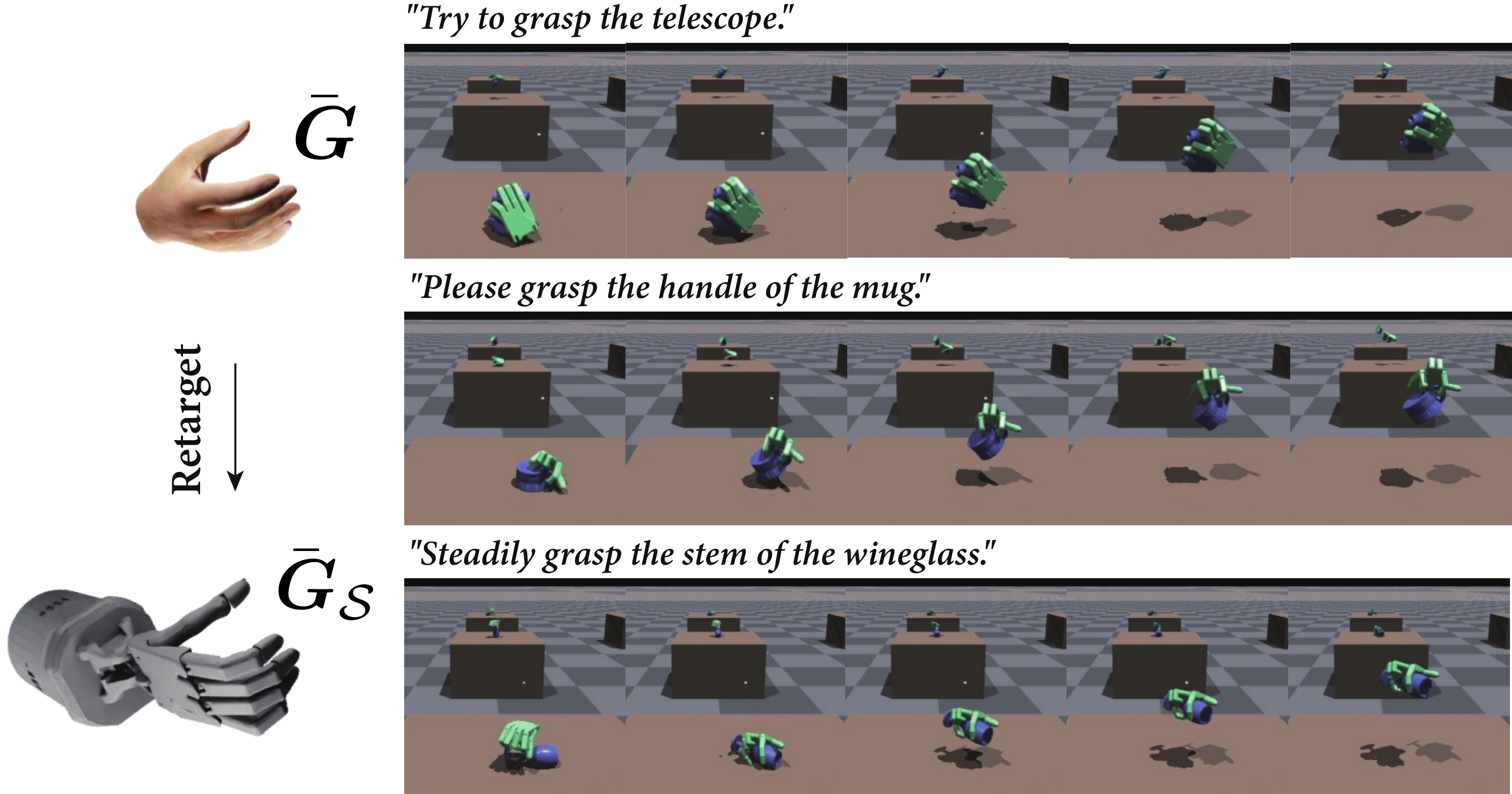}
    \caption{\textbf{Application in robotics.} We verify our results by retargeting the generated grasp to the dexterous ShadowHand. The left image displays the adaptation outcomes, while the right image illustrates the grasp execution process.}
    \label{fig:supp_shadow}
\end{figure}

Given the distinct morphological features and DoFs between the human hand and the ShadowHand, we devise a specialized pipeline for adapting the generated grasp $\bar{\grasp}$ to the ShadowHand model $\bar{\grasp}_\mathcal{S}$. We manually select several corresponding keypoints on both the MANO and ShadowHand models, with a particular emphasis on the fingertips. For each MANO-based grasp $\bar{\grasp}$, the corresponding ShadowHand grasp $\bar{\grasp}_\mathcal{S}$ is optimized by aligning these selected keypoints. To mitigate issues such as unnatural finger movements and potential finger collisions, we introduce a loss function that imposes angular constraints on the ShadowHand's joints, thereby promoting physically plausible adaptations. The outcome of this fitting process is illustrated in \cref{fig:supp_shadow} left. Following this adaptation, the refined grasp $\bar{\grasp}_\mathcal{S}$ is executed using UniDexGrasp's pretrained policy. To enhance the fidelity of collision detection, object meshes are preprocessed using Manifoldplus \cite{huang2020manifoldplus}, followed by convex decomposition algorithms \cite{wei2022approximate}. The results of these grasp executions are displayed in \cref{fig:supp_shadow} right, showcasing the practicality and effectiveness of our methodology in the field of embodied robotics.

\section{\dataset collection}
\cheading{Prompts} As mentioned in our main paper, we craft a set of prompts to direct both GPT-4 and GPT-4v in generating high-quality annotations automatically. For high-level details concerning manipulation intent and grasp status, two specialized prompts are utilized. These are detailed in \cref{table:supp_prompt_contact_annotation,table:supp_prompt_image_annotation}, designed to annotate high-level insights based on the contact information and images respectively.
Additionally, to foster the generation of conversational content, another prompt is crafted to steer GPT-4 in creating conversation templates. This specific prompt, intended to enrich our dataset with conversation annotations, is outlined in \cref{table:supp_prompt_conversation}.\del{Through these carefully designed prompts, we aim to ensure that the annotations not only cover the technical aspects of the grasps but also embody the nuanced, real-world contexts in which these grasps could occur, thus providing a comprehensive and richly detailed dataset.}

\begin{figure}[tb]
    \centering
    \includegraphics[width=\linewidth]{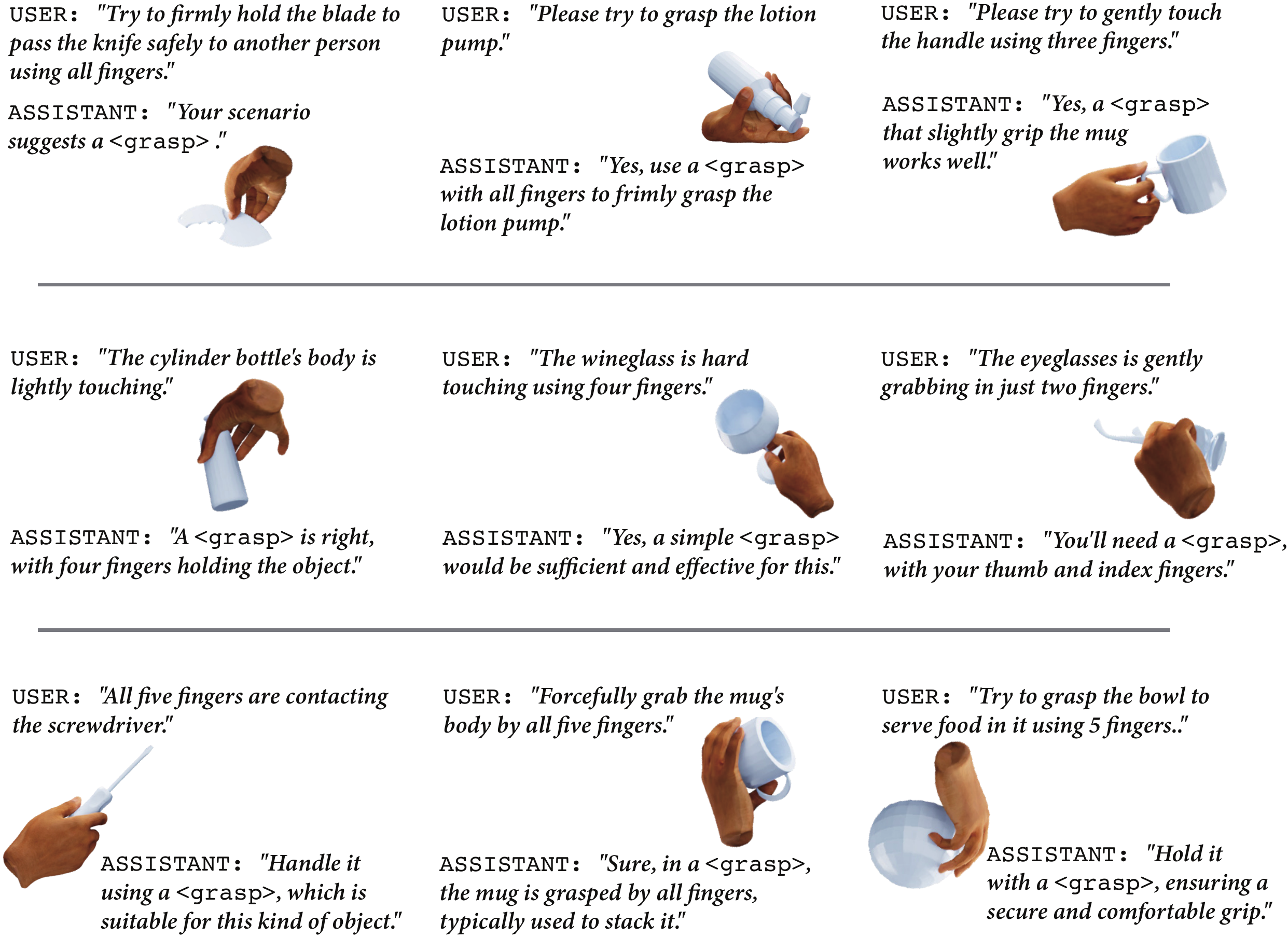}
    \caption{Visualization of our \dataset dataset.}
    \label{fig:supp_dataset_vis}
\end{figure}

\cheading{Dataset Visualization} 
Our \dataset dataset is showcased in \cref{fig:supp_dataset_vis}, featuring an array of objects spanning various categories, shapes, and functionalities. Accompanying language instructions encompass a broad spectrum of grasp intentions and object interactions, with each object annotated with at least 8-10 distinct manipulation intents. This rich diversity is essential for the training of our \method, enabling the generation of grasps that are not only semantically coherent but also physically plausible across diverse scenarios.

\begin{table}[t]
    \begin{center}
        \begin{minipage}[!htp]{1.0\linewidth}
            \begin{center}
                \caption{System Prompt of MLLM.}
                \label{table:supp_prompt_mllm}
                \vspace{-1.2em}
                \resizebox{1.0\linewidth}{!}{
                    \begin{tcolorbox} 
                        \centering
                        \scriptsize
                        \begin{itemize}[leftmargin=0mm]
                        \setlength{\itemsep}{1pt}
                        \item A chat between a curious user and an artificial intelligence assistant. The assistant gives helpful, detailed, and polite answers to the user's questions.  The assistant can understand the information of the three-dimensional object model provided by the user, and combine the knowledge of human hand grasping to assist the user. The following is the object model information: $\obj$. 
                        \end{itemize}
                    \end{tcolorbox}
            }
            \vspace{0.2em}
            \end{center}
            \end{minipage}
            \;
        \begin{minipage}[!htp]{0.455\linewidth}
        \begin{center}
            \caption{Prompt of evaluation.}
            \label{table:supp_prompt_eval}
            \vspace{-0.8em}
            \resizebox{1.0\linewidth}{!}{
                \begin{tcolorbox} 
                    \centering
                    \scriptsize
                    \begin{itemize}[leftmargin=0mm]
                    \setlength{\itemsep}{1pt}
                    \item Your task is to evaluate the alignment between a hand pose and its corresponding textual description, focusing on the grasp's intent, specific object parts engaged by th e hand, and the contact dynamics between fingers and the object. Disregard the background and any textures on both the hand and object. Assessments should consider the physical feasibility of the grasp. Scores will range from 0 to 100, where 100 signifies perfect alignment and a physically plausible grasp, while 0 indicates a significant misalignment or a grasp that seriously defies physical principles. Just directly give the final score, such as: {95}. Here is the description $\lang$ and the grasp image $\mbold{I}$:
                    \end{itemize}
                \end{tcolorbox}
        }
        \vspace{0.5em}
        \end{center}
        \end{minipage}
        \;
        \begin{minipage}[!htp]{0.52\linewidth}
        \begin{center}
            \caption{Prompt of contact based on high-level annotations.}
            \label{table:supp_prompt_contact_annotation}
            \vspace{-1.5em}
            \resizebox{1.0\linewidth}{!} {
                \begin{tcolorbox} 
                    \centering
                    \scriptsize
                    \begin{itemize}[leftmargin=2mm]
                    \setlength{\itemsep}{1pt}
                    \item Given a formatted input describing which part of an object is grasped by a human hand and how many fingers are contacting the object, your task is to generate a set of grasp purposes or add more grasping details, such as the grasping method, the grasping force, etc. Please provide a comprehensive list (8 - 10 phrases) of potential grasp purposes or additional grasping details for each object and its corresponding grasped part.

                    \item For example:
                    \#input: [\{OBJ: "mug", PART: "handle", FINGERS: 5\}, \{OBJ: "pen", PART: "", FINGERS: 3\}]
                    \#output: [\{"to drink", "to make a toast"\}, \{"to write", "to underline or highlight text"\}]
                    Please note that some PART may be empty, indicating that the entire object is grasped.
                    \end{itemize}
                \end{tcolorbox}
            }
        \end{center}
        \vspace{0.5em}
    \end{minipage}
    \begin{minipage}[!htp]{0.40\linewidth}
        \begin{center}
            \caption{Prompt of image-based high-level annotations.}
            \label{table:supp_prompt_image_annotation}
            \vspace{-1.em}
            \resizebox{1.0\linewidth}{!}{
                \begin{tcolorbox} 
                    \centering
                    \scriptsize
                    \begin{itemize}[leftmargin=0mm]
                    \setlength{\itemsep}{1pt}
                    \item Deduce the manipulation intent and the grasp force status from an image depicting hand-object interaction. Your response should focus on the object's contact part, affordance, hand grasp types, and hand fingers status to support your answer. Please ignore the background and object texture when deducing. Provide a clear and concise answer of the manipulation intent and grasp force status, following the example: \{"intent": "to cut something", "status": "firmly grasping"\}. Your analysis should be thorough and accurate, considering all relevant aspects of the hand-object interaction to support your deductions effectively.
                    \end{itemize}
                \end{tcolorbox}
        }
        \end{center}
        \end{minipage}
        \;
        \begin{minipage}[!htp]{0.57\linewidth}
            \begin{center}
                \caption{Prompt of conversation templates generation.}
                \label{table:supp_prompt_conversation}
                \vspace{-1.em}
                \resizebox{1.0\linewidth}{!}{
                    \begin{tcolorbox} 
                        \centering
                        \scriptsize
                        \begin{itemize}[leftmargin=0mm]
                        \setlength{\itemsep}{1pt}
                        \item Please provide conversation templates related to the topic of "human grasping the object." The conversation should incorporate the optional elements: 1. \{finger\}: e.g. "four fingers", "only one finger" 2. \{status\}: e.g. "firmly contacting", "softly touching" 3. \{intent\}: e.g. "pour water", "cut something", "toast", "transfer food onto a plate" 4. \{object\}: e.g. "mug", "bottle" Along with the required element of <grasp> as a state noun. Build one round of conversation using these elements, allowing for flexibility and creativity in the conversation templates. You should focus on creating dialogue that reflects human interaction related to grasping objects, considering various scenarios and details provided by the optional elements. Here is some examples: \{"USER": "Try to \{status\} the \{object\} using \{finger\}". "ASSISTANT": "Sure, here is the <grasp>."\} \{"USER": "\{finger\} are \{status\} the \{object\} \{intent\}". "ASSISTANT": "This sounds like a typical <grasp>."\}
                        \end{itemize}
                    \end{tcolorbox}
            }
            \end{center}
            \end{minipage}
            \;
    \end{center}
\end{table}

\end{appendix}

\end{document}